\documentclass{article}

 \usepackage[preprint]{neurips_2026}

\usepackage[utf8]{inputenc} 
\usepackage[T1]{fontenc}    
\usepackage{url}            
\usepackage{booktabs}       
\usepackage{amsfonts}       
\usepackage{nicefrac}       
\usepackage{microtype}      
\usepackage{xcolor}         

\usepackage{amsmath}
\usepackage{amssymb}
\usepackage{mathtools}
\usepackage{amsthm}
\usepackage{algorithm}
\usepackage{algorithmic}
\usepackage{graphicx}
\usepackage{caption}
\usepackage{subcaption}

\usepackage{adjustbox}

\usepackage{hyperref}       

\theoremstyle{plain}
\newtheoremstyle{TheoremNum}
    {\topsep}{\topsep}              
    {\itshape}                      
    {}                              
    {\bfseries}                     
    {.}                             
    { }                             
    {\thmname{#1}\thmnote{ \bfseries #3}}
\theoremstyle{TheoremNum}
\newtheorem{proposition-repeat}{Proposition}

\title{SPROUT: Predicting Objective-Grounded Prompt Utility for Efficient Text-to-Image Preference Tuning}

\author{%
  Kangping Hu \\
  Georgia Institute of Technology\\
  \texttt{kangping.hu@gatech.edu} \\
  \AND
  Akshay Mehra \\
  Dolby \\
  \texttt{akshay.mehra@dolby.com} \\
  \And
  Stephen Mussmann \\
  Georgia Institute of Technology\\
  \texttt{mussmann@gatech.edu} \\
}

\begin{document}

\maketitle

\begin{abstract}
Preference tuning of text-to-image diffusion models spends most of its compute generating candidate images, yet training prompts are typically sampled uniformly at random, wasting compute generating images that provide too little signal for model alignment. 
To best utilize the available compute, we study prompt selection, where at each iteration we identify the most useful prompts for preference tuning.
We separate prompt selection into a utility score, an estimator of that utility score, and batch selection. 
Within this framework, we propose Selection by PRedicting Objective-grounded UTility (SPROUT). 
The utility score is predicted on every prompt in the pool, including unseen prompts, by a text-only model, refit after every iteration on all utilities observed so far. 
A randomized acceptance rule then turns these predictions into a batch, so no images are generated beyond those used for training.
For direct group preference optimization (DGPO), the objective-grounded utility is the norm of the gradient that the objective produces from a prompt's group of images.
Fine-tuning Stable Diffusion 1.5 with DGPO on GenEval, SPROUT matches the final score of uniform sampling in about 44\% of the iterations and improves the area under the learning curve with respect to both uniform sampling and CGPO, a recent variance-based prompt sampler. 
The efficiency gains hold under four held-out reward models never used in training, and the ranking of methods carries over to DPG-Bench and T2I-CompBench++.
Our code is available at \url{https://github.com/ADDAPT-ML/SPROUT}.
\end{abstract}

\section{Introduction}
\label{sec:intro}

Preference tuning has become a standard approach to align generative models with human
judgment, including text-to-image (T2I) diffusion models, where we collect comparisons between generated images and tune the model toward the preferred
ones~\citep{rafailov2023dpo,wallace2024diffusiondpo}. The comparisons can come from human annotators, from learned reward models, or from hand-crafted reward pipelines such as GenEval~\citep{ghosh2023geneval}, which makes the procedure cheap enough to iterate.

While assigning reward scores has become more efficient, generating images to be scored remains expensive. Aligning a diffusion model requires generating
multiple images per prompt, scoring them, and computing a gradient update on the scored generated images, so the cost of an iteration is linear in the number of prompts: doubling the prompts per iteration doubles the time and computation of preference tuning. Generating and scoring the images also takes more than half of an iteration in our runs (about 56\%, Appendix~\ref{sec:impl}), so the cost of a prompt is avoided only if the prompt is discarded before generation.

This generation cost motivates prompt selection: not all prompts are equally worth generating from.
A prompt
whose candidates come out uniformly strong, or uniformly weak, yields a poorly separated comparison and
contributes little, yet consumes the same generation budget as one that yields a well-separated group.
Identifying the useful prompts \emph{before} generating can reduce the computational cost directly. We call this the
\textbf{prompt selection} problem.

Active learning is one precedent~\citep{mackay1992information,lewis1994sequential,sener2018active,kirsch2019batchbald},
but the resource it conserves is \emph{annotation} of data that already exists. Here the data do not exist
until they are paid for, and the resource is \emph{generation}, so a prompt selection algorithm has to predict
the value of a sample it has not yet seen. Closer to our setting, prompt curricula in reinforcement learning select or filter prompts before or during generation~\citep{yu2025dapo,zheng2025greso,foster2025learnability,Li_2026_CVPR,adagrpo2026}.
The closest to our setting, CGPO~\citep{Li_2026_CVPR}, samples text-to-image prompts by a prompt-specific estimate of the reward variance, which cannot distinguish among prompts that have not been generated from, the vast majority of a large prompt pool. More generally, a reward statistic is not directly tied to how the preference objective uses a prompt's images~\citep{gao2025pcl,qu2026gps,qu2025mopps,yu2025dapo,jiang2025vcrl,zheng2025greso,Li_2026_CVPR}, and probe generations spend the budget that selection is meant to save~\citep{yu2025dapo,jiang2025vcrl,adagrpo2026}.

\begin{figure*}[t]
  \centering
  \includegraphics[width=\linewidth]{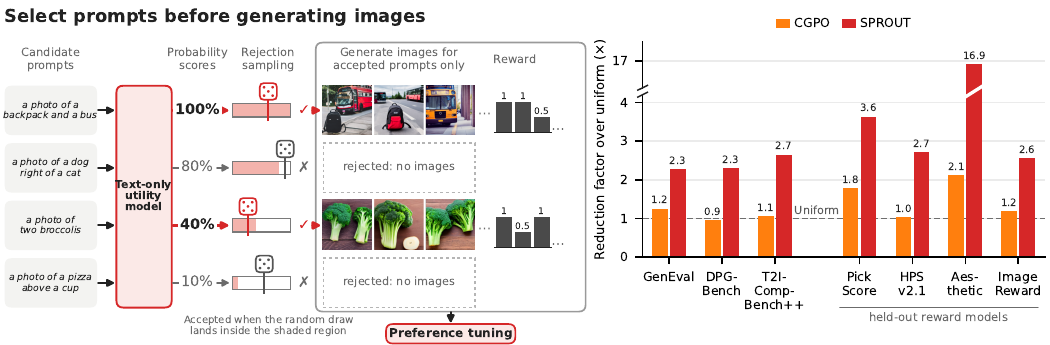}
  \caption{{\bf Left.} SPROUT accepts each prompt at random with a probability scaled from its predicted utility, the update norm, so a rejected prompt generates no images (details in Figure~\ref{fig:main}). {\bf Right.} The reduction factor of each method, i.e., the factor by which it reduces the training iterations needed to reach uniform sampling's best score, computed on the mean curve of three seeds; for T2I-CompBench++, on the mean of its eight categories as no official overall score is defined.}
  \label{fig:overview}
\end{figure*}

In this work, we propose \textbf{SPROUT} (\textbf{S}election by \textbf{PR}edicting \textbf{O}bjective-grounded \textbf{UT}ility), which selects prompts by what the preference objective does with them rather than by a statistic of the reward. After each training iteration, every selected prompt receives a label: the norm of the gradient that the objective produces from its group of images, a by-product of the training step that costs no extra generation. A text-only \emph{utility model}, refit online on all labels observed so far, generalizes these labels to every prompt in the pool, including prompts never generated from. Its predictions are then scaled to be acceptance probabilities, so that the next batch favors high-utility prompts while no prompts are completely excluded (Figure~\ref{fig:overview}).

SPROUT is not tied to a particular objective: a prompt's label comes from the gradient of the training loss, so it is defined for any preference objective whose loss is a sum over prompt groups, which includes the group-relative objectives used for diffusion models, and it needs no design specific to the reward.

Our contributions are:
\begin{itemize}
    \item A text-only utility model, refit online, that estimates the utility of every prompt in the pool from its text.
    \item An objective-grounded utility: the norm of the per-prompt gradient produced by the preference objective.
    \item An evaluation on Stable Diffusion 1.5 tuned with DGPO on GenEval~\citep{ghosh2023geneval}, where our method reduces the training iterations needed to reach uniform sampling's final score by a factor of $2.27$ (the \emph{reduction factor}, which measures the training that a given score requires), improves the area under the learning curve over uniform sampling and over CGPO~\citep{Li_2026_CVPR} at no extra generation cost, and keeps its advantage under four held-out reward models~\citep{kirstain2023pickapic,wu2023hpsv2, schuhmann2022laion5b,schuhmann2022aesthetic,xu2023imagereward} and on DPG-Bench~\citep{hu2024ella} and T2I-CompBench++~\citep{huang2025t2icompbench}.
\end{itemize}

\section{Related work}

\textbf{Preference optimization tuning.} DPO~\citep{rafailov2023dpo} removes the explicit reward model of preference tuning with PPO~\citep{schulman2017ppo,ziegler2019finetuning,ouyang2022instructgpt}, and Diffusion-DPO~\citep{wallace2024diffusiondpo} adapts it to diffusion. Reward-backpropagation methods differentiate through sampling~\citep{clark2024draft,prabhudesai2024alignprop}, DDPO~\citep{black2024ddpo} applies reinforcement learning to diffusion, and GRPO~\citep{shao2024grpo}, Dr.GRPO~\citep{liu2025drgrpo}, Flow-GRPO~\citep{liu2025flowgrpo}, DanceGRPO~\citep{xue2025dancegrpo} and DGPO~\citep{luo2026reinforcing} compare rewards within a per-prompt group, across reasoning and visual generation.
However, they all assume a prompt distribution is given or use uniform prompt selection, and none of them asks which prompts to spend generation time and compute on.

\textbf{Prompt curricula and rollout allocation.} For language-model reasoning, DAPO~\citep{yu2025dapo} filters out prompts whose rollouts are all correct or all incorrect, GRESO~\citep{zheng2025greso} skips prompts with a zero-variance streak, and LILO~\citep{foster2025learnability} estimates the learnability $p(1-p)$ of candidate prompts from a few rollouts each, where $p$ is the success rate of the current model on the prompt, and trains on the most learnable ones.
PCL~\citep{gao2025pcl}, GPS~\citep{qu2026gps}, and MoPPS~\citep{qu2025mopps} predict prompt difficulty, Knapsack RL~\citep{li2025knapsack} studies rollout-budget allocation, and LearnAlign~\citep{li2026learnalign} selects a training subset after warm-up by gradient alignment weighted by success rate, which requires rollouts for every candidate.

\begin{figure*}[t]
  \centering
  \includegraphics[width=\linewidth]{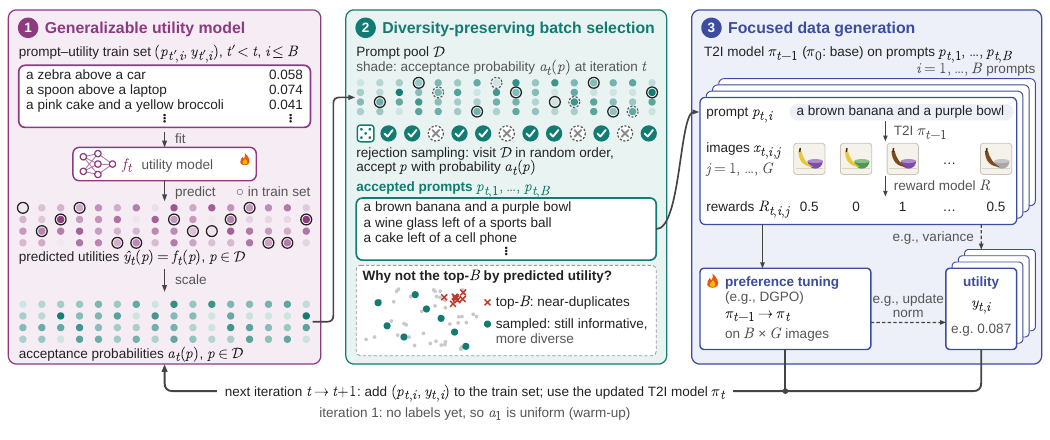}
   \caption{Overview of SPROUT at iteration $t$, with stages 1 and 2 as in Algorithm~\ref{alg:select}. (1) Generalizable utility model: a text-only utility model $f_t$ is fit on the train set of prompt--utility pairs from earlier iterations (circled prompts) and predicts the utility $\hat{y}_t(p)$ of every prompt in $\mathcal{D}$, including prompts never generated from; its predictions are scaled into acceptance probabilities $a_t(p)$. (2) Diversity-preserving batch selection: the pool $\mathcal{D}$ is visited in a random order and each prompt is accepted with probability $a_t(p)$ until $B$ prompts $p_{t,1},\dots,p_{t,B}$ are accepted; the inset contrasts this with top-$B$ selection (Appendix~\ref{sec:ablation}). (3) Focused data generation: $\pi_{t-1}$ generates $G$ images for each accepted prompt only, the reward model scores them, and preference tuning updates the model to $\pi_t$; each accepted prompt $p_{t,i}$ is then labeled with its utility $y_{t,i}$, computed from its rewards or from the update (dashed arrows), and $(p_{t,i}, y_{t,i})$ joins the train set of the next iteration.}
   \label{fig:main}
\end{figure*}

\textbf{Text-to-image prompt selection.} FiFA~\citep{yang2025fifa} filters an offline human-feedback dataset by preference margin, text quality, and diversity, and UniFL~\citep{zhang2024unifl} selects informative and diverse prompts once before training, with a semantic filter and nearest-neighbor compression rather than a learned score. These methods are offline and the prompt selection does not evolve with the growing model capabilities. AdaGRPO~\citep{adagrpo2026} generates one probe image per candidate prompt and selects the candidate whose probe reward is closest to a capability anchor, a moving average of the rewards the model has recently obtained. However, this requires additional generations for selection that are not used for training. The work most similar to ours is CGPO~\citep{Li_2026_CVPR}, which updates historical variance-based sampling probabilities from the training generations only but loses the opportunity to intelligently select among prompts that have not been generated from. SPROUT is online, requires no additional generations, and can also extrapolate to ungenerated prompts.

\begin{algorithm}[h]
\caption{Online prompt selection for text-to-image preference optimization using SPROUT.}
\label{alg:workflow}
\begin{algorithmic}[1]
\REQUIRE prompt pool $\mathcal{D}$, budget $B$, group size $G$, iterations $T$, reward function $R$.
\STATE $\pi_0 \gets$ base model, $\mathcal{O}_0 \gets \emptyset$
\FOR{$t = 1 \ldots T$}
  \STATE // \textit{Select $B$ text prompts based on historical outputs}
  \STATE $\{p_{t,i}\}_{i=1}^{B} \gets \textsc{SPROUT}(\mathcal{D}, B, \{\mathcal{O}_{t'}\}_{t'<t})$
  \FOR{prompts $i=1,\dots,B$}
  \FOR{images $j=1,\dots,G$}
  \STATE // \textit{Generate j$^{th}$ image from prompt $i$}
  \STATE $x_{t,i,j} \sim \pi_{t-1}(p_{t,i})$
  \STATE // \textit{Score generated image relative to prompt}
  \STATE $R_{t,i,j} \gets R(x_{t,i,j},p_{t,i})$
  \ENDFOR
  \ENDFOR
  \STATE // \textit{Update text-to-image model with new data}
  \STATE $\pi_t \gets \textsc{Train}(\pi_{t-1}, \{(p_{t,i},x_{t,i,j},R_{t,i,j})\}_{i,j})$
  \STATE // \textit{Save outputs (incl. gradients) for future selection}
  \STATE $\mathcal{O}_t \gets \{(p_{t,i}, \{x_{t,i,j}\}_{j=1}^{G}, \{R_{t,i,j}\}_{j=1}^{G}, g_{t,i})\}_{i=1}^{B}$ 
  \ENDFOR
\RETURN $\pi_T$
\end{algorithmic}
\end{algorithm}

\textbf{Learned utility models.} MoPPS~\citep{qu2025mopps} performs streaming Bayesian inference on each prompt's success rate to avoid the rollout cost, a per-prompt estimate that does not generalize across prompts. GPS~\citep{qu2026gps} trains a small generative predictor on the shared optimization history and combines predicted difficulty with batch diversity, and PCL~\citep{gao2025pcl} learns a value model of prompt success to hold a curriculum near intermediate difficulty. However, they are all in the LLM reasoning domain and use difficulty as the utility only. SPROUT instead uses the update the preference objective makes from a prompt as our learned utility, and to the best of our knowledge, we are the first to use a learned utility model for prompt selection in text-to-image preference tuning.

\textbf{Data selection.} Active learning aims to determine what to label, where selecting
a batch to annotate can be as tricky as our batch prompt selection. Batch-aware Bayesian methods
address the redundancy that arises when top-scoring items are selected together~\citep{kirsch2019batchbald,hu2025batch}, while stochastic batch
selection recovers much of that benefit by perturbing scores before top scoring~\citep{kirsch2023stochastic}.
Besides, Coreset selection~\citep{sener2018active}, prototypical multi-batch $k$-means~\citep{pourahmadi2021simple},
and data pruning~\citep{sorscher2022beyond} motivate coverage-based selection. Per-example gradient norms, or bounds on them, have also ranked examples for data pruning~\citep{paul2021datadiet} and importance sampling~\citep{katharopoulos2018notall}, and BADGE~\citep{ash2020badge} combines gradient magnitude with batch diversity. These methods require access to the unlabeled data, whereas our setting requires selection \emph{before} generating the unlabeled data.

\section{Preliminaries}
\label{sec:preliminaries}

\subsection{Setting}

Let $\mathcal{D}$ be the pool of training prompts. A text-to-image model is trained over $T$ iterations with a per-iteration budget of $B \ll |\mathcal{D}|$ prompts, where $\pi_t$ denotes the text-to-image model after iteration $t$, with trainable parameters $\theta$, so that $\pi_0$ is the base model. At iteration $t$ (Algorithm~\ref{alg:workflow}), a \emph{selection strategy} chooses prompts $p_{t,i} \in \mathcal{D}$ for $i=1,\dots,B$, the model $\pi_{t-1}$ generates $G$ images $x_{t,i,j}\sim\pi_{t-1}(p_{t,i})$, $j=1,\dots,G$, for each selected prompt, and a reward model $R$ assigns rewards $R_{t,i,j}=R(x_{t,i,j},p_{t,i})$ for each generated image. $\pi_{t-1}$ is then updated to $\pi_t$ by training on the prompts, images, and rewards with a preference optimization objective (see Section~\ref{sec:objective}).
We evaluate the model throughout learning curves on a held-out test set of prompts $\mathcal{Q}$ disjoint from $\mathcal{D}$, reporting the mean score $\frac{1}{|\mathcal{Q}|}\sum_{q \in \mathcal{Q}} R'(x_q, q)$ under a judge $R'$: a benchmark's reward model or a held-out reward model.
We report the score after $T$ iterations and the area under the learning curve, as well as the reduction factor in the number of iterations needed to reach a baseline's best score.

\subsection{Preference tuning for text-to-image models}
\label{sec:objective}

We consider group-relative preference objectives in the style of GRPO~\citep{shao2024grpo}, such as Flow-GRPO~\citep{liu2025flowgrpo} and DGPO~\citep{luo2026reinforcing} for diffusion models, which sample a group of images per prompt and optimize a loss on rewards standardized within the group. Given the group of $G$ images of $p_{t,i}$, with reward mean $\bar{R}_{t,i}$ and standard deviation $\mathrm{sd}_{t,i}$, each image is standardized to the advantage $A_{t,i,j} = (R_{t,i,j} - \bar{R}_{t,i})/\mathrm{sd}_{t,i}$, and the loss at iteration $t$ is a sum of per-group terms,
\begin{equation}
\mathcal{L}_t=\frac{1}{B}\sum_{i=1}^{B}\big[\ell_{t,i}(\theta)+\lambda\,r_{t,i}(\theta)\big],
\label{eq:grouploss}
\end{equation}
where $\ell_{t,i}$ is the loss of the objective on the group of $p_{t,i}$, computed from its images $x_{t,i,j}$ and advantages $A_{t,i,j}$, and $r_{t,i}$, with weight $\lambda$, a penalty toward the reference model on the same images, such as a KL penalty. We write $g_{t,i}$ for the gradient of $\mathcal{L}_t$ contributed by the group of $p_{t,i}$, so that the batch gradient is $g_t=\sum_{i=1}^{B}g_{t,i}$. A group whose rewards are all equal ($A_{t,i,j}\equiv 0$) contributes only through the penalty. Prompt selection decides which groups enter this sum, and the utility of SPROUT (Section~\ref{sec:target}) needs only the per-group gradient $g_{t,i}$, which every loss of this form provides.

\textbf{DGPO.} We instantiate Equation~\ref{eq:grouploss} with direct group preference optimization (DGPO)~\citep{luo2026reinforcing}. We choose DGPO because it is a recent group-relative objective for diffusion models, and its authors report that it improves the GenEval score with far less training than Flow-GRPO~\citep{liu2025flowgrpo}. Let $\mathcal{E}^{\theta}(x,p) = \|\epsilon_\theta(x_\tau, \tau, p) - \epsilon\|^2$ be the denoising
error of image $x$ under policy $\theta$ at a sampled timestep $\tau$ and noise $\epsilon$, and
$\mathcal{E}^{\text{ref}}$ for the same quantity under the frozen base model. Each group contributes
\begin{align}
&S_{t,i} = \frac{1}{G}\sum_{j=1}^{G} A_{t,i,j}\big[\, \mathcal{E}^{\text{ref}}(x_{t,i,j},p_{t,i}) - \mathcal{E}^{\theta}(x_{t,i,j},p_{t,i}) \,\big],
\\
&\mathcal{L}_{\text{DGPO},t} = \frac{1}{B}\sum_{i=1}^{B}\frac{1}{\beta}\mathrm{softplus}(-\beta S_{t,i}),
\label{eq:dgpo}
\end{align}
where $\mathrm{softplus}(z)=\ln(1+e^{z})$, or equivalently $\mathrm{softplus}(-z) = -\ln\sigma(z)$, with $\sigma$ the logistic function. Intuitively, $S_{t,i}$ is the advantage-weighted margin by which the tuned model improves on the reference model for
positively-advantaged samples relative to negatively-advantaged ones, and $\beta$ is a scaling factor. One $(\tau, \epsilon)$ draw is
shared across a group, so that all of its samples are compared under the same corruption noise.
The gradient of Equation~\ref{eq:dgpo} is $-\frac{1}{B}\sum_{i=1}^{B}w_{t,i}\,\nabla_\theta S_{t,i}$ where each group's contrast is scaled by the logistic weight $w_{t,i}=\sigma(-\beta S_{t,i})$. The per-group loss of Equation~\ref{eq:grouploss} is thus $\ell_{t,i}=\frac{1}{\beta}\mathrm{softplus}(-\beta S_{t,i})$, and the penalty $r_{t,i}$ is a KL penalty to the reference model, following the official DGPO implementation.

\subsection{Online prompt selection}
\label{sec:online}

We abstract online prompt selection into three separate components: a \emph{utility}, the prompt-level quantity that selection should favor; an \emph{estimator}, which scores candidate prompts by that utility; and \emph{batch selection}, which uses the scores to select the $B$ prompts of an iteration. Existing methods correspond to particular choices of the three, which we review below; SPROUT (Section~\ref{sec:method}) contributes a new utility and a new estimator.

\subsubsection{Utility}

\textbf{Difficulty}~\citep{gao2025pcl,qu2026gps,qu2025mopps}. The \emph{difficulty}
of generating from a prompt $p$ can be defined as the negative expected training reward for $p$ at iteration $t$:
$\mu_{t}(p)$ $= -\mathbb{E}_{x \sim \pi_{t-1}(p)}[R(x,p)]$.
However, the standardization of the advantages (Section~\ref{sec:objective}) removes the group's mean reward and its raw scale from the weights, so a prompt's absolute difficulty does not enter the update magnitude directly, though prior work uses it as a heuristic for selection.

\textbf{Variance}~\citep{yu2025dapo,jiang2025vcrl,zheng2025greso,Li_2026_CVPR}. Since the group objective requires multiple images for each prompt, another prompt utility
can be the \emph{variance} of their rewards across different generated images for each prompt $p$:
$\sigma^2_{t}(p)$ $= \mathbb{E}_{x \sim \pi_{t-1}(p)}\left[\left(R(x,p) - \mathbb{E}_{x \sim \pi_{t-1}(p)}[R(x,p)]\right)^2\right]$
as the variance utility for $p$ at iteration $t$.
A group with zero variance has zero advantages in group-based preference tuning, which is why existing work either discards zero-variance prompts~\citep{yu2025dapo,zheng2025greso} or favors high-variance prompts~\citep{jiang2025vcrl,Li_2026_CVPR}.

\subsubsection{Estimator}

Existing estimators either generate fresh images to measure the utility or keep a separate estimate for each prompt from the groups already trained on.

\textbf{Oracle probe}~\citep{yu2025dapo,jiang2025vcrl,adagrpo2026} generates $n$ fresh samples per candidate prompt $p$ and ranks by realized difficulty
$\mu_{t}(p)$, variance
$\sigma^2_{t}(p)$, or other utilities.
It measures the utility directly, but it requires an adequately large $n$ to be accurate and spends $n$ generations on every candidate it scores, including the candidates it then discards. Scoring the whole pool once would cost $n\times|\mathcal{D}|$ images, several times the $T\times B\times G$ images of an entire training run in our setting, which is why the methods in this category score only a sampled set of candidates, and most of them are for LLM reasoning only.

\textbf{Prompt-specific estimator}~\citep{zheng2025greso,Li_2026_CVPR,qu2025mopps} shares nothing across prompts. It maintains a per-prompt estimate,
updated from the realized utility whenever that prompt is generated from, and leaves every never-observed
prompt at its initial value or naively updates its probability. It can learn to avoid prompts it has paid for,
but has no evidence about a prompt it has not generated from, so it can treat such a prompt only as it treats every other unmeasured one.

\subsubsection{Batch selection}

\emph{Batch selection} uses the scores to select a batch of $B$ prompts.

\textbf{Top-$B$}~\citep{gao2025pcl,qu2025mopps}, the default in most selection work, takes the $B$ highest scores
; it is deterministic given the scores and concentrates the batch on a narrow band of the score range, discarding prompt variety that a preference optimization update may need.

\textbf{Rejection sampling}~\citep{zheng2025greso,Li_2026_CVPR} visits the pool in a random order and accepts each prompt $p$ with an acceptance probability $a_t(p)\in[0,1]$ computed from its score until $B$ are accepted; it keeps the batch broad while favoring high-utility prompts.

\section{SPROUT}
\label{sec:method}

SPROUT instantiates the three components of Section~\ref{sec:online} with the update norm as a new utility (Section~\ref{sec:target}), a text-only utility model refit online on all labels observed so far as a new estimator (Section~\ref{sec:estimator}), and rejection sampling as the batch selection (Section~\ref{sec:strategies}), as shown in Figure~\ref{fig:main} and Algorithms~\ref{alg:workflow} and~\ref{alg:select}.

\begin{algorithm}[t]
\caption{\textsc{SPROUT}$(\mathcal{D}, B, \{\mathcal{O}_{t'}\}_{t'<t})$ at iteration $t$. Numbers refer to the stages of Figure~\ref{fig:main}; $c_{\max}$ and $a_{\min}$ are given in Table~\ref{tab:impl}.
}
\label{alg:select}
\begin{algorithmic}[1]
\REQUIRE pool $\mathcal{D}$, budget $B$, outputs $\{\mathcal{O}_{t'}\}_{t'<t}$ (Algorithm~\ref{alg:workflow}), threshold $c_{\max}$, floor $a_{\min}$
\STATE // \textit{Stage 1: label, fit the utility model and scale its predictions.}
\IF{$t=1$}
  \STATE $a_{t}(p)\gets 1$ for all $p\in\mathcal{D}$ \COMMENT{Initialization}
\ELSE
  \STATE $y_{t',i}\gets u_{t',i}$ (Eq.~\ref{eq:updatenorm}) for $t'<t$, $i\leq B$ \COMMENT{Label}
  \STATE fit $f_{t}$ on $\{(p_{t',i}, y_{t',i})\}_{t'<t,\,i\leq B}$ \COMMENT{Fit}
  \STATE $\hat{y}_{t}(p)\gets f_{t}(p)$ for all $p\in\mathcal{D}$ \COMMENT{Estimate}
  \STATE $c(p)\gets\min(\hat{y}_{t}(p),\,$$c_{\max}$$)$ for all $p\in\mathcal{D}$ \COMMENT{Truncate}
  \STATE $a_{t}(p)\gets\max\!\big(\frac{c(p)-\min_q c(q)}{\max_q c(q)-\min_q c(q)},\,$$a_{\min}$$\big)$ \COMMENT{Scale}
\ENDIF
\STATE // \textit{Stage 2: select $B$ prompts.}
\STATE $i\gets 0$
\FOR[Rejection sampling]{$p$ in a random order of $\mathcal{D}$}
  \STATE with probability $a_{t}(p)$: $i\gets i+1$, $p_{t,i}\gets p$
  \STATE \textbf{if} $i=B$ \textbf{then break}
\ENDFOR
\RETURN $\{p_{t,i}\}_{i=1}^{B}$
\end{algorithmic}
\end{algorithm}

\begin{figure*}[t]
\centering
\includegraphics[width=\linewidth]{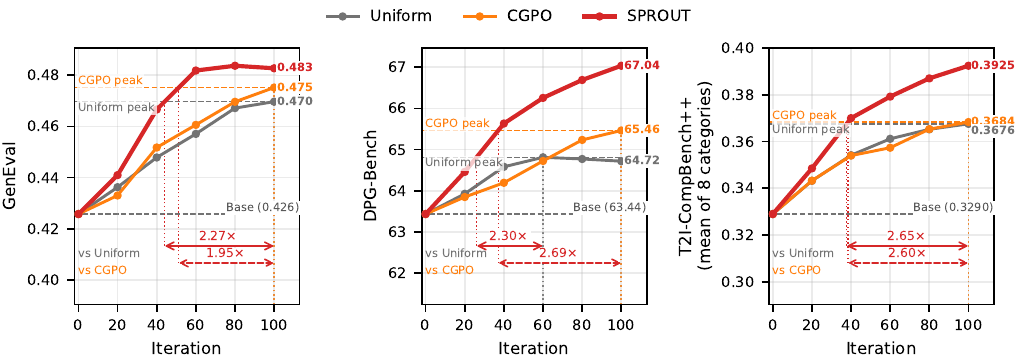}
\caption{Learning curves with $B=8$ on GenEval (left), DPG-Bench (middle) and T2I-CompBench++ (mean of the eight categories, right), mean of three seeds. Arrows mark the reduction factors relative to the peaks of uniform sampling (solid) and CGPO (dashed). SPROUT reduces the iterations needed for uniform sampling's best score by $2.27\times$, $2.30\times$ and $2.65\times$, and lies above both baselines at every evaluated iteration on not only GenEval, but also DPG-Bench and T2I-CompBench++, which no model is trained on.}
\label{fig:curves}
\end{figure*}

\subsection{Objective-grounded utility}
\label{sec:target}

The difficulty and variance of Section~\ref{sec:online} are both functions of the reward, and so are agnostic to the optimizer that actually updates the model. Once a group has been scored and trained on, the objective offers its own per-prompt quantity at no extra cost: the size of the gradient that the group produces. We define the \emph{update norm}
\begin{equation}
u_{t,i}=\log\frac{\lVert g_{t,i}\rVert}{\operatorname{median}_{k\le B}\lVert g_{t,k}\rVert},
\label{eq:updatenorm}
\end{equation}
where $g_{t,i}$ is the gradient of the training loss with respect to the trainable parameters accumulated from prompt $p_{t,i}$'s group at iteration $t$, before gradient clipping. Normalizing by the batch median keeps labels from different iterations on a common scale, and the logarithm treats half and twice the median norm symmetrically, so that a few large gradients do not dominate the regression fit. The ranking by update norm carries over to the optimizer's actual step: the gradient norm and the norm of the AdamW-scaled gradient rank prompts almost identically (mean within-iteration Spearman $0.93$), and the gradient norm also ranks the first-order decrease in each prompt's sampled objective under the actual AdamW step (Spearman $0.86$; details in Appendix~\ref{sec:impl}).

\subsection{Generalizable utility model}
\label{sec:estimator}

SPROUT estimates the utility with a \emph{utility model}, as PCL~\citep{gao2025pcl} and GPS~\citep{qu2026gps} do for difficulty in language-model reasoning: a surrogate model that fits a text-only regressor on records $(p_{t',i}, y_{t',i})$ from previous iterations $t'<t$, where the label $y_{t',i}$ is the realized difficulty, variance, update norm, or other utility of $p_{t',i}$, and ranks each pool prompt $p$, including prompts never generated from, by its predicted utility $\hat{y}_{t}(p)$, spending no generation at
selection time. Each label is computed from the group generated when that prompt was selected, the same quantity the oracle probe measures, but without generating before selection. The utility model is therefore a cost-effective proxy for the utility. The model is refit after each iteration on every record so far; Appendix~\ref{sec:ablation} compares other label histories.
At the first iteration, when no labels exist, every prompt is accepted (the initialization in Algorithm~\ref{alg:select}).

\subsection{Diversity-preserving batch selection}
\label{sec:strategies}

SPROUT selects the batch by rejection sampling (Section~\ref{sec:online}), with the acceptance probability $a_t(p)$ obtained from the prediction $\hat{y}_t(p)$ by truncating, min-max normalizing over the pool and flooring, following CGPO~\citep{Li_2026_CVPR} (Algorithm~\ref{alg:select} and Appendix~\ref{sec:impl}). Fed the same predictions, top-$B$ concentrates the batch on far fewer distinct prompts and is unstable across label histories (Appendix~\ref{sec:ablation}).

\section{Experiments}
\label{sec:experiments}

\subsection{Data and reward models}
\label{sec:data}

To compare the empirical performance of prompt selection methods, we perform preference optimization with the GenEval prompts and reward, and evaluate the trained models via GenEval, held-out reward models, and other benchmarks.
Training prompts come from the GenEval training set released with DGPO~\citep{luo2026reinforcing}, following the official implementation of Flow-GRPO~\citep{liu2025flowgrpo} ($50{,}000$ prompts), and the training reward is a partial-credit GenEval reward computed from the official detector's per-requirement verdicts, inspired by Flow-GRPO~\citep{liu2025flowgrpo}, DGPO~\citep{luo2026reinforcing}, and CGPO~\citep{Li_2026_CVPR}. Evaluation follows the official GenEval~\citep{ghosh2023geneval} protocol: $553$ prompts with four images each, macro-averaging the fraction of correct images over the six task categories. We also re-score the same evaluation images with four held-out reward models that never enter training: PickScore~\citep{kirstain2023pickapic}, HPS v2.1~\citep{wu2023hpsv2}, the LAION aesthetic predictor~\citep{schuhmann2022laion5b,schuhmann2022aesthetic} and ImageReward~\citep{xu2023imagereward}. Following CGPO, we also evaluate the GenEval-trained checkpoints on DPG-Bench~\citep{hu2024ella}, and on T2I-CompBench++~\citep{huang2025t2icompbench}.

\begin{figure*}[t]
\centering
\includegraphics[width=0.7\linewidth]{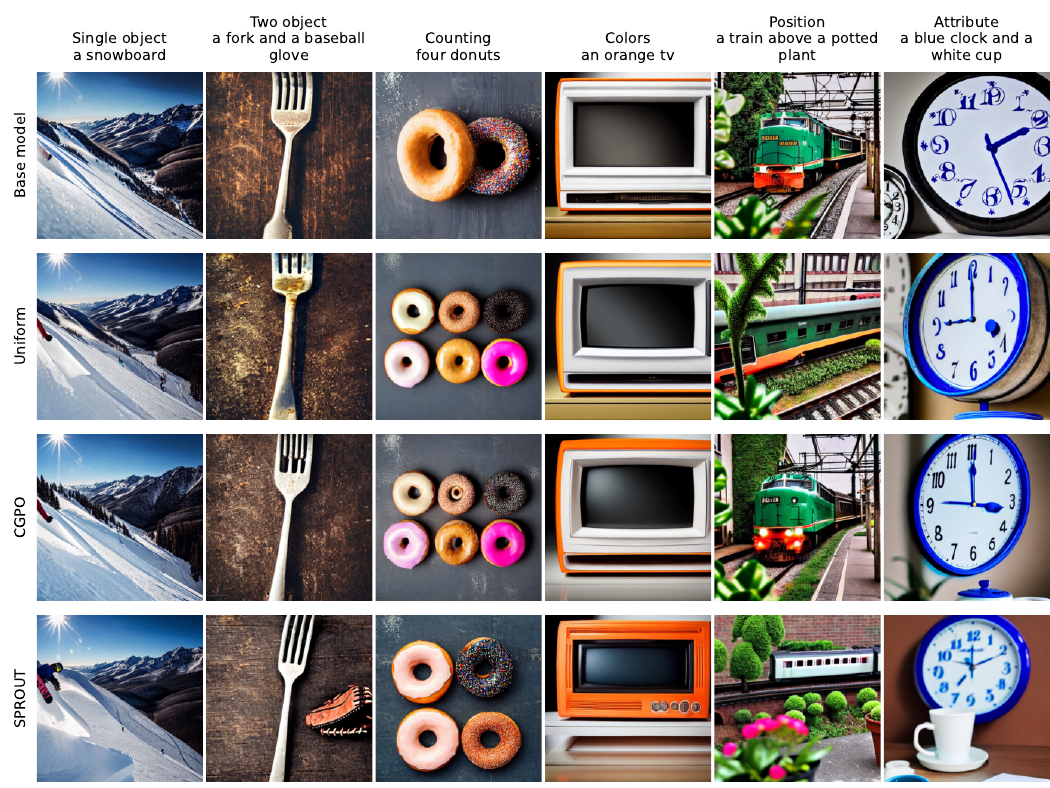}
\caption{Qualitative comparison on one selected GenEval test prompt per task category after 100 iterations, with the same prompt and sampling seed in each column. SPROUT follows the prompts more accurately than the base model, uniform sampling and CGPO: it generates every requested object with the correct count, color, spatial relation and attribute binding.
}
\label{fig:qualitative}
\end{figure*}

\subsection{Prompt selection algorithms}
\label{sec:algorithms}

We compare against uniform sampling, the reference every selection strategy is measured against, and CGPO~\citep{Li_2026_CVPR}, a state-of-the-art text-to-image prompt-selection method.
The appendix ablates each component of SPROUT: the utility, replaced by the reward variance (SPROUT w/ variance, Appendix~\ref{sec:utilitydiff}), which existing work uses to discard zero-variance prompts~\citep{yu2025dapo,zheng2025greso} or to favor high-variance ones~\citep{jiang2025vcrl,Li_2026_CVPR}; the label history and batch selection (Appendix~\ref{sec:ablation}); and the predictor (Appendix~\ref{sec:predabl}).

\subsection{Training configuration and statistics}
\label{sec:model}

\textbf{Text-to-image models.} We focus on low-budget, low-computation scenarios, which are often overlooked in preference tuning. We use Stable Diffusion 1.5~\citep{rombach2022ldm}, a latent diffusion model, tuned with a LoRA~\citep{hu2022lora} adapter. We follow DGPO's released GenEval configuration, adapted to the Stable Diffusion 1.5 backbone, with a constant learning rate of $1.65\times10^{-4}$ selected by a hyperparameter search with uniform sampling in preliminary experiments without using the test set. We train for $T=100$ iterations, each with $B=8$ prompts and $G=24$ images per prompt, for a total of 19{,}200 generated training images. More details are in Appendix~\ref{sec:impl}.

\textbf{Utility model.} Each pool prompt is embedded once with the \emph{frozen} text encoder of OpenAI CLIP ViT-L/14~\citep{radford2021clip} ($768$ dimensions, $\ell_2$-normalized). After every iteration, a ridge regression with penalty $\alpha=10$ is refit on the update norms of all prompts trained on so far and predicts every prompt in the pool.

\begin{figure*}[t]
\centering
\includegraphics[width=0.9\linewidth]{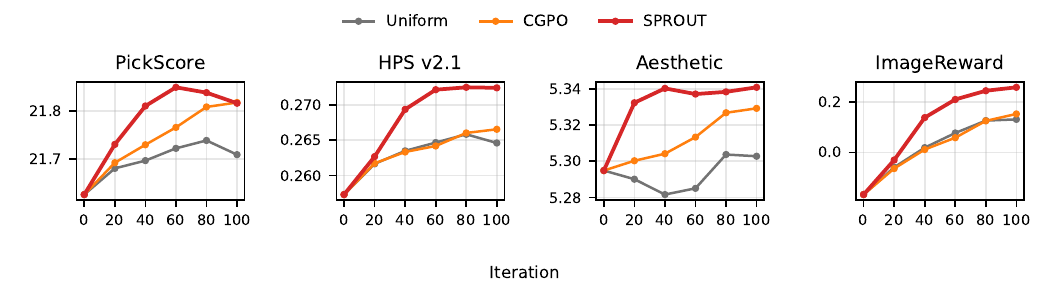}
\caption{Held-out reward models over training, mean of three seeds; iteration $0$ is the base model. The gain is not specific to the detector behind the training reward: SPROUT improves nAULC over uniform sampling under all four reward models, each with an interval that excludes zero (Table~\ref{tab:judges}).}
\label{fig:judges}
\end{figure*}

\textbf{Statistics.} Every method is trained with the same three seeds, fixed before any of the reported experiments, and we report means over the three seeds with ``$\pm$'' the half-width of the 95\% confidence interval from the Student's $t$ distribution. Differences between methods are paired on seed and reported as the mean with its 95\% $t$ interval in brackets; bold marks intervals that exclude zero. The normalized area under the learning curve (nAULC) is the area under a benchmark's learning curve, evaluated every $20$ iterations from the base model (iteration $0$) to iteration $100$, divided by $100$. The reduction factor is the iteration at which a baseline's mean curve peaks divided by the iteration at which a method's linearly interpolated mean curve first reaches that score, i.e., the factor by which the method reduces the training iterations needed for the baseline's best score. Applied to each seed separately, SPROUT reaches uniform sampling's best score on every seed, under every benchmark and held-out reward model (Appendix Table~\ref{tab:perseedrf}).

\subsection{Results}
\label{sec:results}

\textbf{Main results.} Figure~\ref{fig:overview} summarizes the reduction factors across benchmarks, Figure~\ref{fig:curves}~(left, see Table~\ref{tab:main} in the appendix for exact numbers) compares the methods on GenEval, and Table~\ref{tab:genevalcat} in the appendix breaks the comparison down by task category, with example generations in Figure~\ref{fig:qualitative}. SPROUT reaches uniform sampling's final GenEval score in less than half the iterations (a reduction factor of $2.27\times$) and its nAULC gain is significant against both uniform sampling ($+0.0143$) and CGPO ($+0.0124$). CGPO, in contrast, improves on uniform sampling only slightly, with an interval that includes zero: its per-prompt estimates are informed only by the prompts it has generated from, a small fraction of the pool when $B\times T\ll|\mathcal{D}|$, whereas our utility model predicts the utility of every prompt at no extra generation cost. Figure~\ref{fig:informative} in Appendix~\ref{sec:informative} shows examples of prompts that the utility model ranks highest and lowest: images of a highest-ranked prompt receive more varied rewards than images of a lowest-ranked one, providing more useful training signal to preference optimization.

\textbf{Held-out reward models.} Figure~\ref{fig:judges} (exact numbers in Table~\ref{tab:judges} of the appendix) shows that the gain is not specific to the object detector that the training reward and the GenEval metric share: SPROUT improves nAULC over uniform sampling under all four, with every interval excluding zero, while the intervals of CGPO include zero under all four.

\textbf{Transfer to DPG-Bench.} Figure~\ref{fig:curves}, middle, demonstrates that the same ordering holds on DPG-Bench, which no model is trained on (exact numbers in Tables~\ref{tab:main} and~\ref{tab:dpgcat} in the appendix): SPROUT lies above uniform sampling and CGPO in nAULC on every seed, significantly so against CGPO.

\textbf{Transfer to T2I-CompBench++.} On T2I-CompBench++, SPROUT improves on uniform sampling and CGPO in seven of the eight categories after 100 iterations, and in each of them it is ahead of both baselines on every seed. Figure~\ref{fig:curves}, right, shows the mean of the eight categories, and SPROUT has a statistically significant gain in nAULC against both CGPO and uniform sampling (exact numbers in Tables~\ref{tab:main} and~\ref{tab:compbench} of the appendix).

\textbf{Ablations.} With the reward variance as the utility, the GenEval nAULC gain over uniform sampling remains significant but drops to $+0.0097$, and the gain over CGPO has an interval that includes zero (Appendix~\ref{sec:utilitydiff}). Selecting the top $B$ predictions instead of sampling concentrates the batch on far fewer distinct prompts, lowers the mean gain of SPROUT and, with the reward variance as the utility, collapses on one seed (Appendix~\ref{sec:ablation}). Predictors that rank prompts better offline do not improve training with $B=16$ (Appendix~\ref{sec:predabl}).

\textbf{Larger budgets.} Doubling the batch to $B=16$ keeps training stable, and SPROUT stays above uniform sampling in nAULC on every seed (Appendix~\ref{sec:batch16}). Extended to 200 iterations with a decaying learning rate, SPROUT keeps its lead on the seed mean under GenEval, the held-out reward models, DPG-Bench and T2I-CompBench++ (Appendix~\ref{sec:beyond}). In both settings, however, the interval of SPROUT's GenEval gain over uniform sampling includes zero.

\section{Conclusion and limitations}
\label{sec:conclusion}

We frame prompt selection for preference tuning of text-to-image models as three separate components --- utility, estimator, and batch selection --- and propose SPROUT, whose objective-grounded utility is the norm of the gradient that the preference objective produces from a prompt's group. The utility is predicted from prompt text by a text-only model refit online, and prompts are accepted with probabilities scaled from the predicted utility. SPROUT reaches uniform sampling's final GenEval score in less than half the iterations, improves the area under the learning curve over uniform sampling and CGPO at no extra generation cost, and lies above both baselines in nAULC under four held-out reward models and on DPG-Bench and T2I-CompBench++, which it is not trained on.

\textbf{Limitations.} Our experiments use a single backbone (Stable Diffusion 1.5), preference objective (DGPO) and training reward (the partial-credit GenEval reward). In the low-budget setting we focus on, the update norm gives a larger gain in GenEval nAULC and reduction factor than the reward variance, although the interval of the nAULC difference includes zero. The reward variance may work better with larger budgets (Appendix~\ref{sec:utilitydiff}), and with $B=16$ and over 200 iterations the GenEval gains hold on the seed mean while their intervals include zero (Appendices~\ref{sec:batch16} and~\ref{sec:beyond}). Predictors that rank prompts better offline did not always improve training (Appendix~\ref{sec:predabl}), and prompt selection beyond 100 iterations requires a decaying learning rate to stay stable (Appendix~\ref{sec:beyond}). We leave these directions for future work.

\bibliography{reference}
\bibliographystyle{plainnat}


\clearpage
\appendix

{\centering
\Large
\textbf{Appendix}\\
\vspace{1.0em}}

\noindent\textbf{Contents.}
\begin{itemize}\setlength{\itemsep}{0pt}
\item \hyperref[sec:impl]{Appendix~\ref*{sec:impl}\hspace{1em}\nameref*{sec:impl}}
\item \hyperref[sec:overall]{Appendix~\ref*{sec:overall}\hspace{1em}\nameref*{sec:overall}}
\item \hyperref[sec:utilitydiff]{Appendix~\ref*{sec:utilitydiff}\hspace{1em}\nameref*{sec:utilitydiff}}
\item \hyperref[sec:catdelta]{Appendix~\ref*{sec:catdelta}\hspace{1em}\nameref*{sec:catdelta}}
\item \hyperref[sec:informative]{Appendix~\ref*{sec:informative}\hspace{1em}\nameref*{sec:informative}}
\item \hyperref[sec:ablation]{Appendix~\ref*{sec:ablation}\hspace{1em}\nameref*{sec:ablation}}
\item \hyperref[sec:batch16]{Appendix~\ref*{sec:batch16}\hspace{1em}\nameref*{sec:batch16}}
\item \hyperref[sec:predabl]{Appendix~\ref*{sec:predabl}\hspace{1em}\nameref*{sec:predabl}}
\item \hyperref[sec:beyond]{Appendix~\ref*{sec:beyond}\hspace{1em}\nameref*{sec:beyond}}
\end{itemize}

\section{Implementation details}
\label{sec:impl}

\begin{table}[h]
\centering
\small
\caption{Training configuration shared by every method on GenEval in our main experiments, with the rejection sampling values used by the methods that sample.}
\label{tab:impl}
\begin{tabular}{@{}ll@{}}
\toprule
Setting & Value \\
\midrule
Backbone & Stable Diffusion 1.5 \\
Adapter & LoRA, rank $64$, $\alpha=128$ \\
Optimizer & AdamW, $\beta_2=0.999$ \\
Updates per iteration & $1$ \\
Learning rate & $1.65\times10^{-4}$, constant \\
Prompts per iteration & $B=8$ \\
Images per prompt & $G=24$ \\
Iterations & $100$ \\
Rollouts & $10$ DDIM steps, guidance $4.5$ \\
KL coefficient $\lambda$ & $0.02$ \\
$\beta$ & $100$ \\
Clip range & $0.01$ \\
Truncation threshold & $c_{\max}=0.6$ \\
Acceptance floor & $a_{\min}=0.1$ \\
Training reward & GenEval, partial credit \\
Evaluation weights & EMA, decay $0.9$ \\
\bottomrule
\end{tabular}
\end{table}

\textbf{Training.} Table~\ref{tab:impl} lists the configuration. All numeric values other than the learning rate (Section~\ref{sec:model}) and the rejection sampling values, which follow CGPO, follow DGPO's released GenEval configuration. Every method uses the same configuration and differs only in how the $B$ prompts of each iteration are chosen. The training reward of an image is the fraction of the prompt's requirements that the official GenEval verifier finds satisfied, with each requirement evaluated separately by the official GenEval code.

\textbf{Rejection sampling.} Predictions are truncated at the threshold $c_{\max}$, min-max normalized over the pool and floored at $a_{\min}$, following CGPO's released code. The prompts of the pool are visited in a random order and each is accepted with its probability; the sweep repeats until $B$ distinct prompts are accepted. The random order depends only on the seed and the iteration.

\textbf{CGPO.} We implement CGPO following its released code: a prompt-specific estimate of the reward variance, calibrated per GenEval task category as in the paper, turned into a batch by the same rejection sampling. CGPO's other techniques belong to its prompt-specific estimator: every prompt not measured in a round has its probability raised by an exploration increment, and each probability is averaged with its two previous values. The category calibration is the only place where task labels enter training and is only used by CGPO.

\textbf{Update-norm label.} The gradient $g_{t,i}$ in Equation~\ref{eq:updatenorm} is that of the full per-sample loss of the official DGPO code, the policy term plus the KL penalty, accumulated over the $G$ images of $p_{t,i}$ and the training timesteps in fp32 before gradient clipping. A sample clipped by the trust region, or a group whose rewards are all equal, contributes only through the KL penalty. At iteration 1 the policy equals the reference model, so such a group has a zero gradient; the ratio in Equation~\ref{eq:updatenorm} is therefore floored at $0.0199$, the smallest ratio observed in a preliminary run, before taking the logarithm.

\textbf{Optimizer diagnostics.} The correlations in Section~\ref{sec:target} come from two continuations of one SPROUT run (seed A; for simplicity, we call the three training seeds A, B and C, whose values are given in the released code), from iterations 40 and 80 for 20 iterations each with unchanged settings, which log each prompt's gradient $g$ together with AdamW's diagonal scaling $D$ and the realized step $\Delta\theta$, the new parameters minus the old. Within each iteration, among the $298$ groups whose rewards are not all equal, we compute the Spearman correlation of $\lVert g\rVert$ with $\lVert Dg\rVert$ (mean $0.93$, standard deviation $0.07$ over the $40$ iterations) and with the first-order decrease $-\langle g,\Delta\theta\rangle$ of the prompt's sampled objective, on the same images and noise ($0.86$, standard deviation $0.11$). The decrease is positive for every one of these groups. The most direct per-prompt quantity of the objective is the logistic weight $w_{t,i}=\sigma(-\beta S_{t,i})$ that scales each group's gradient (Section~\ref{sec:objective}); in our measured runs, however, it is nearly constant across prompts, so it cannot rank them, and what separates prompts is the size of the gradient their group produces. The cross terms between prompts are also small in our runs: over iterations 2--100 of the three SPROUT runs, the ratio $\lVert\sum_{i}g_{t,i}\rVert^2/\sum_{i}\lVert g_{t,i}\rVert^2$, which is $1$ for orthogonal gradients and $B$ for aligned ones, has mean $1.01$ (10th to 90th percentile over iterations $0.96$--$1.07$; norm-weighted mean pairwise cosine $0.0015$), so $\lVert\sum_{i}g_{t,i}\rVert^2\approx\sum_{i}\lVert g_{t,i}\rVert^2$, and the norm of a group's gradient approximates its share of the batch update.

\textbf{Evaluation.} GenEval uses the official $553$ prompts with four images each, $50$ sampling steps, guidance $9$ and seed $42$, every $20$ iterations, scored with the official detector. The pipeline's default safety checker stays on, as in the official code. The held-out reward models re-score these images with their authors' released code: PickScore (the README's inference code), HPS v2.1 (the \texttt{hpsv2} package), the LAION aesthetic predictor (\texttt{simple\_inference.py}, CLIP ViT-L/14 with the \texttt{sac+logos+ava1-l14-linearMSE} head) and ImageReward (the \texttt{image-reward} package). DPG-Bench uses the official harness at the same checkpoints, with four $512\times512$ images per prompt and the mPLUG-large VQA model. The release does not specify sampler settings, so every checkpoint, the base model included, uses the pipeline defaults ($50$ steps, guidance $7.5$) with seed $42+k$ for the $k$-th image. T2I-CompBench++ uses the $300$ validation prompts of each category at the same checkpoints, with ten images per prompt and the settings of the official generation script (DDIM, $30$ steps, guidance $7.5$, seeds $42$ to $51$, the pipeline's default safety checker). The scorers are the released BLIP-VQA (Color, Shape, Texture), UniDet (2D-Spatial, 3D-Spatial, Numeracy), CLIPScore (Non-Spatial) and 3-in-1 (Complex) code. The runs of Appendices~\ref{sec:batch16} and~\ref{sec:beyond} are scored on DPG-Bench and T2I-CompBench++ only at their last iteration, 100 and 200 respectively.

\textbf{Compute.} Each run uses one GPU (NVIDIA Quadro RTX 6000 or V100 32GB). An iteration with $B=8$ takes about $607$ seconds, so a 100-iteration run takes about $17$ GPU-hours; with $B=16$ an iteration takes about $1{,}204$ seconds. Under uniform sampling, generating and scoring the images takes $56\%$ of an iteration, and the gradient update with checkpoint saving takes the remaining $44\%$ (medians over $295$ iterations, measured from the modification times of the files that each stage writes). Prompt selection only adds $5$--$7\%$ to the time per iteration, since its labels come from the training step and its predictions from one regression over the pool.

\begin{table*}[t]
\centering
\small
\caption{Results after 100 iterations (Stable Diffusion 1.5, DGPO, $B=8$, $G=24$), with every model trained on GenEval and evaluated every 20 iterations. Mean $\pm$ 95\% interval over three seeds; differences are paired on seed, with bold intervals excluding zero.}
\label{tab:main}
\begin{subtable}{\linewidth}
\centering
\caption{GenEval, with differences $\times10^{-3}$.}
\begin{adjustbox}{max width=\linewidth}\begin{tabular}{lccccc}
\toprule
Method & GenEval@100 & nAULC & $\Delta$nAULC vs Uniform & $\Delta$nAULC vs CGPO & Reduction factor \\
 & & & ($\times10^{-3}$) & ($\times10^{-3}$) & \\
\midrule
Base model & 0.426 & -- & -- & -- & -- \\
Uniform & 0.470\,$\pm$\,0.051 & 0.451\,$\pm$\,0.013 & -- & -- & 1.00$\times$ \\
CGPO~\cite{Li_2026_CVPR} & 0.475\,$\pm$\,0.015 & 0.453\,$\pm$\,0.012 & +1.9\,[-9.4, +13.2] & -- & 1.25$\times$ \\
SPROUT & 0.483\,$\pm$\,0.026 & 0.465\,$\pm$\,0.010 & \textbf{+14.3\,[+5.8, +22.7]} & \textbf{+12.4\,[+9.5, +15.3]} & 2.27$\times$ \\
\bottomrule
\end{tabular}
\end{adjustbox}
\end{subtable}
\par\medskip
\begin{subtable}{\linewidth}
\centering
\caption{DPG-Bench, with differences in DPG-Bench points.}
\begin{adjustbox}{max width=\linewidth}\begin{tabular}{lccccc}
\toprule
Method & DPG-Bench@100 & nAULC & $\Delta$nAULC vs Uniform & $\Delta$nAULC vs CGPO & Reduction factor \\
\midrule
Base model & 63.44 & -- & -- & -- & -- \\
Uniform & 64.72\,$\pm$\,4.46 & 64.44\,$\pm$\,1.53 & -- & -- & 1.00$\times$ \\
CGPO~\cite{Li_2026_CVPR} & 65.46\,$\pm$\,1.61 & 64.49\,$\pm$\,0.96 & +0.06\,[-1.33, +1.44] & -- & 0.94$\times$ \\
SPROUT & 67.04\,$\pm$\,0.55 & 65.65\,$\pm$\,0.92 & +1.22\,[-0.35, +2.79] & \textbf{+1.16\,[+0.96, +1.37]} & 2.30$\times$ \\
\bottomrule
\end{tabular}
\end{adjustbox}
\end{subtable}
\par\medskip
\begin{subtable}{\linewidth}
\centering
\caption{T2I-CompBench++, mean of the eight categories, with differences $\times10^{-3}$.}
\begin{adjustbox}{max width=\linewidth}\begin{tabular}{lccccc}
\toprule
Method & T2I-CompBench++@100 & nAULC & $\Delta$nAULC vs Uniform & $\Delta$nAULC vs CGPO & Reduction factor \\
\midrule
Base model & 0.3290 & -- & -- & -- & -- \\
Uniform & 0.3676\,$\pm$\,0.0381 & 0.3544\,$\pm$\,0.0169 & -- & -- & 1.00$\times$ \\
CGPO~\cite{Li_2026_CVPR} & 0.3684\,$\pm$\,0.0203 & 0.3537\,$\pm$\,0.0119 & -0.7\,[-13.2, +11.8] & -- & 1.05$\times$ \\
SPROUT & 0.3925\,$\pm$\,0.0115 & 0.3691\,$\pm$\,0.0045 & \textbf{+14.7\,[+0.4, +29.0]} & \textbf{+15.4\,[+8.1, +22.7]} & 2.65$\times$ \\
\bottomrule
\end{tabular}
\end{adjustbox}
\end{subtable}
\caption{Held-out reward models: difference in nAULC from uniform sampling (PickScore, Aesthetic and ImageReward $\times10^{-2}$; HPS v2.1 $\times10^{-3}$).}
\label{tab:judges}
\begin{adjustbox}{max width=\linewidth}\begin{tabular}{lcccc}
\toprule
Method & PickScore & HPS v2.1 & Aesthetic & ImageReward \\
\midrule
CGPO~\cite{Li_2026_CVPR} & +4.3\,[-0.1, +8.6] & +0.1\,[-2.3, +2.5] & +1.9\,[-1.2, +5.1] & -0.4\,[-7.9, +7.1] \\
SPROUT & \textbf{+8.9\,[+4.6, +13.1]} & \textbf{+5.0\,[+4.7, +5.3]} & \textbf{+4.1\,[+2.1, +6.1]} & \textbf{+9.3\,[+4.4, +14.3]} \\
\bottomrule
\end{tabular}
\end{adjustbox}
\end{table*}

\section{Overall results}
\label{sec:overall}

\begin{figure*}[t]
\centering
\includegraphics[width=\linewidth]{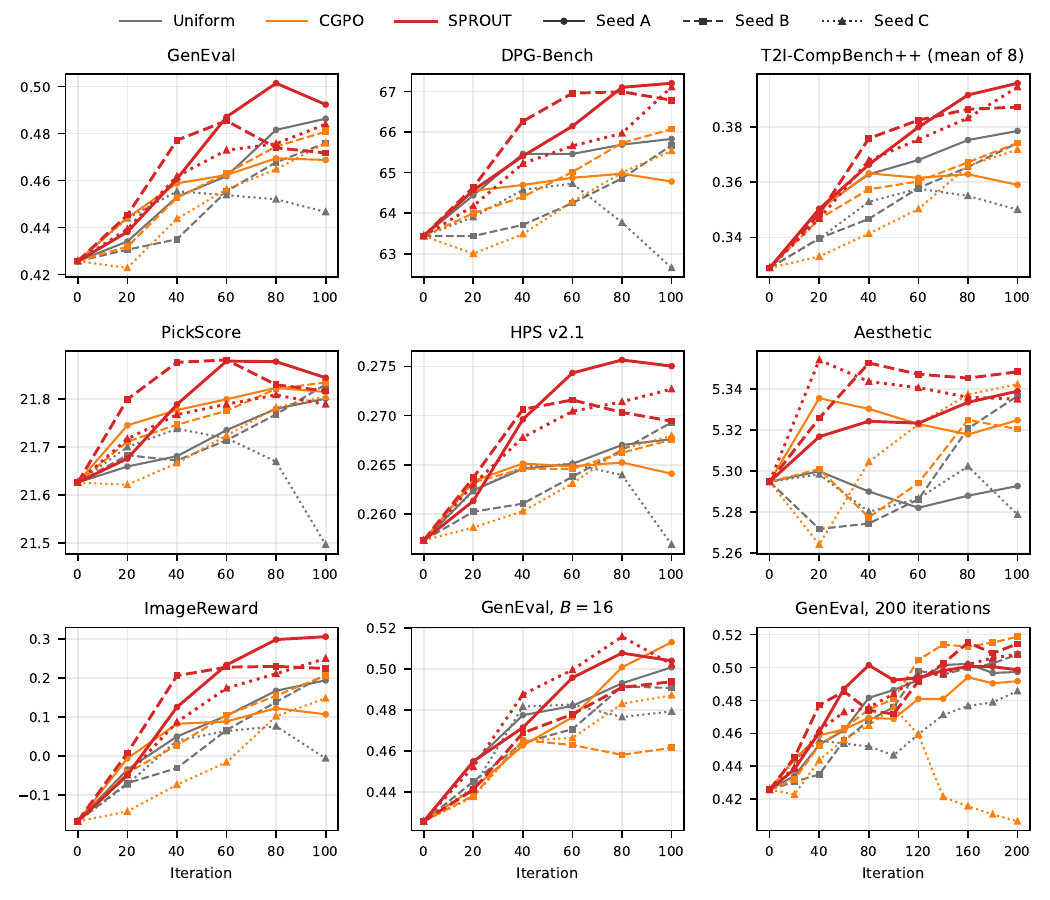}
\caption{Learning curves of each seed, one panel per setting: GenEval, DPG-Bench, T2I-CompBench++ (mean of the eight categories) and the four held-out reward models with $B=8$ over 100 iterations, GenEval with $B=16$ (Appendix~\ref{sec:batch16}), and GenEval over 200 iterations with the learning rate decayed after iteration 100 (Appendix~\ref{sec:beyond}). Color marks the method and line style the seed; iteration $0$ is the base model.}
\label{fig:perseed}
\end{figure*}

Table~\ref{tab:main} gives the scores, nAULC and reduction factors on GenEval, DPG-Bench and T2I-CompBench++, and Table~\ref{tab:judges} gives the differences under the held-out reward models. Table~\ref{tab:perseedrf} gives the reduction factors on each seed and Figure~\ref{fig:perseed} the learning curves of each seed in every setting. SPROUT reaches uniform sampling's best score on all three seeds on every benchmark and under every held-out reward model ($21$ of $21$), CGPO in $13$ of $21$. The factor of a single seed can differ substantially from that of the mean curve: the Aesthetic factor of $16.91\times$ on the mean curve is $4.30$, $3.56$ and $31.73$ on seeds A, B and C, the last because that seed's curve reaches uniform sampling's best score within the first 20 iterations.

\begin{table*}[t]
\centering
\small
\caption{Reduction factor relative to uniform sampling on each seed, i.e., the iteration at which uniform sampling's curve of that seed peaks divided by the interpolated iteration at which the method's curve of the same seed first reaches that score, and on the mean curve of the three seeds as in the main text; -- marks a method that does not reach uniform sampling's best score within $100$ iterations.}
\label{tab:perseedrf}
\begin{adjustbox}{max width=\linewidth}\begin{tabular}{lcccccccc}
\toprule
 & \multicolumn{4}{c}{SPROUT} & \multicolumn{4}{c}{CGPO} \\
\cmidrule(lr){2-5}\cmidrule(lr){6-9}
Benchmark & Seed A & Seed B & Seed C & Mean curve & Seed A & Seed B & Seed C & Mean curve \\
\midrule
GenEval & 1.68 & 2.55 & 1.17 & 2.27 & -- & 1.18 & 0.68 & 1.25 \\
DPG-Bench & 1.94 & 3.05 & 1.96 & 2.30 & -- & 1.28 & 0.83 & 0.94 \\
T2I-CompBench++ & 1.72 & 2.58 & 1.98 & 2.65 & -- & 1.00 & 0.86 & 1.05 \\
PickScore & 2.34 & 3.55 & 1.41 & 3.63 & 1.64 & 1.06 & 0.62 & 1.78 \\
HPS v2.1 & 2.84 & 2.76 & 2.09 & 2.72 & -- & -- & 0.84 & 1.03 \\
Aesthetic & 4.30 & 3.56 & 31.73 & 16.91 & 7.97 & -- & 2.06 & 2.12 \\
ImageReward & 1.90 & 2.33 & 2.08 & 2.55 & -- & -- & 1.06 & 1.19 \\
\bottomrule
\end{tabular}
\end{adjustbox}
\end{table*}

\begin{table*}[t]
\centering
\small
\caption{SPROUT w/ variance and SPROUT against the baselines with $B=8$, paired on seed, with bold intervals excluding zero: GenEval nAULC ($\times10^{-3}$), nAULC under the held-out reward models (scaled as in Table~\ref{tab:judges}), DPG-Bench nAULC (points), and T2I-CompBench++ nAULC of the mean of the eight categories ($\times10^{-3}$).}
\label{tab:variance}
\begin{adjustbox}{max width=\linewidth}\begin{tabular}{llcc}
\toprule
Benchmark & & SPROUT w/ variance & SPROUT \\
\midrule
GenEval & $\Delta$nAULC vs Uniform & \textbf{+9.7\,[+5.7, +13.8]} & \textbf{+14.3\,[+5.8, +22.7]} \\
 & $\Delta$nAULC vs CGPO & +7.8\,[-7.4, +23.1] & \textbf{+12.4\,[+9.5, +15.3]} \\
 & Reduction factor & 1.80$\times$ & 2.27$\times$ \\
\midrule
Held-out reward models & $\Delta$nAULC vs Uniform, PickScore & +5.0\,[-0.7, +10.7] & \textbf{+8.9\,[+4.6, +13.1]} \\
 & $\Delta$nAULC vs Uniform, HPS v2.1 & \textbf{+3.1\,[+2.5, +3.7]} & \textbf{+5.0\,[+4.7, +5.3]} \\
 & $\Delta$nAULC vs Uniform, Aesthetic & \textbf{+3.3\,[+0.4, +6.2]} & \textbf{+4.1\,[+2.1, +6.1]} \\
 & $\Delta$nAULC vs Uniform, ImageReward & \textbf{+6.2\,[+0.8, +11.7]} & \textbf{+9.3\,[+4.4, +14.3]} \\
\midrule
DPG-Bench & $\Delta$nAULC vs Uniform & \textbf{+0.90\,[+0.33, +1.47]} & +1.22\,[-0.35, +2.79] \\
 & Reduction factor & 1.96$\times$ & 2.30$\times$ \\
\midrule
T2I-CompBench++ & $\Delta$nAULC vs Uniform & \textbf{+9.9\,[+4.0, +15.7]} & \textbf{+14.7\,[+0.4, +29.0]} \\
 & Reduction factor & 2.00$\times$ & 2.65$\times$ \\
\bottomrule
\end{tabular}
\end{adjustbox}
\end{table*}

\section{Ablation: reward variance as the utility}
\label{sec:utilitydiff}

SPROUT w/ variance keeps every component of SPROUT and replaces only its utility with the reward variance $\sigma^2_{t}(p_{t,i})$ (Section~\ref{sec:online}), computed from the rewards in $\mathcal{O}_{t}$, the statistic that existing work uses to discard zero-variance prompts~\citep{yu2025dapo,zheng2025greso} or to favor high-variance ones~\citep{jiang2025vcrl,Li_2026_CVPR}. Against the baselines with $B=8$ (Table~\ref{tab:variance}), it also improves GenEval nAULC over uniform sampling with an interval that excludes zero ($+9.7\times10^{-3}$, reduction factor $1.80\times$), and likewise under three of the four held-out reward models, on DPG-Bench ($+0.90$ points), and on T2I-CompBench++ ($+9.9\times10^{-3}$). Its gains are smaller than those of SPROUT on GenEval and under every held-out reward model, and its GenEval gain over CGPO has an interval that includes zero.

Table~\ref{tab:utilitydiff} compares the two utilities directly. With $B=8$, SPROUT leads in GenEval nAULC on every seed and under every held-out reward model on average, significantly so under HPS v2.1. SPROUT w/ variance, in turn, ends slightly higher after 100 iterations ($0.493$ against $0.483$) and leads on average in every column with $B=16$, where only the HPS v2.1 interval excludes zero. The reward variance may work better with larger budgets, while in the low-budget setting we focus on, the update norm gives the larger GenEval nAULC gain, with an interval that includes zero, and SPROUT attains the largest reduction factor.

\begin{table}[!ht]
\centering
\footnotesize
\setlength{\tabcolsep}{4pt}
\caption{SPROUT minus SPROUT w/ variance, paired on seed: nAULC and score at iteration 100 on GenEval ($\times10^{-3}$), nAULC under the held-out reward models (scaled as in Table~\ref{tab:judges}) and on DPG-Bench (points). DPG-Bench is scored only at iteration 100 with $B=16$. An interval that excludes zero but would round to zero at one decimal is shown with two.}
\label{tab:utilitydiff}
\begin{adjustbox}{max width=\linewidth}\begin{tabular}{lcc}
\toprule
 & $B=8$ & $B=16$ \\
\midrule
GenEval nAULC & +4.5\,[-7.9, +17.0] & -1.7\,[-21.6, +18.2] \\
GenEval@100 & -9.8\,[-23.3, +3.6] & -4.8\,[-19.7, +10.2] \\
PickScore & +3.9\,[-6.0, +13.7] & -0.6\,[-3.6, +2.4] \\
HPS v2.1 & \textbf{+1.9\,[+0.9, +2.8]} & \textbf{-1.39\,[-2.75, -0.04]} \\
Aesthetic & +0.8\,[-3.0, +4.6] & -0.1\,[-8.8, +8.5] \\
ImageReward & +3.1\,[-7.2, +13.4] & -3.0\,[-9.4, +3.5] \\
DPG-Bench nAULC & +0.32\,[-0.72, +1.36] & -- \\
\bottomrule
\end{tabular}
\end{adjustbox}
\end{table}

\begin{table*}[t]
\centering
\small
\caption{GenEval results by task category.}
\label{tab:genevalcat}
\begin{subtable}{\linewidth}
\centering
\caption{Score after 100 iterations, mean over three seeds; best results in bold.}
\begin{adjustbox}{max width=\linewidth}\begin{tabular}{lccccccc}
\toprule
Method & Single object & Two object & Counting & Colors & Position & Attribute & Overall \\
\midrule
Base model & 0.96 & 0.39 & 0.37 & 0.77 & 0.03 & 0.04 & 0.43 \\
Uniform & 0.96 & 0.51 & 0.39 & 0.78 & 0.06 & 0.10 & 0.47 \\
CGPO & \textbf{0.98} & 0.50 & \textbf{0.41} & \textbf{0.80} & 0.07 & 0.09 & \textbf{0.48} \\
SPROUT & \textbf{0.98} & \textbf{0.58} & 0.34 & 0.79 & \textbf{0.08} & \textbf{0.12} & \textbf{0.48} \\
\bottomrule
\end{tabular}
\end{adjustbox}
\end{subtable}
\par\medskip
\begin{subtable}{\linewidth}
\centering
\caption{Differences in nAULC ($\times10^{-3}$), paired on seed, with bold intervals excluding zero.}
\begin{adjustbox}{max width=\linewidth}\begin{tabular}{lccc}
\toprule
 & \multicolumn{2}{c}{vs Uniform} & vs CGPO \\
\cmidrule(lr){2-3}\cmidrule(lr){4-4}
Category & CGPO & SPROUT & SPROUT \\
\midrule
Single object & +3.2\,[-12.8, +19.2] & +10.4\,[-6.6, +27.5] & +7.2\,[-2.3, +16.6] \\
Two object & -9.6\,[-88.1, +68.9] & \textbf{+47.7\,[+2.2, +93.3]} & \textbf{+57.3\,[+2.0, +112.6]} \\
Counting & +10.4\,[-2.9, +23.7] & -3.4\,[-23.8, +16.9] & \textbf{-13.9\,[-21.1, -6.6]} \\
Colors & \textbf{+5.6\,[+4.9, +6.2]} & +2.1\,[-14.5, +18.8] & -3.5\,[-19.7, +12.8] \\
Position & +2.6\,[-33.3, +38.5] & +15.0\,[-9.4, +39.4] & +12.4\,[-5.2, +30.0] \\
Attribute & -0.9\,[-14.1, +12.3] & +13.8\,[-12.7, +40.2] & \textbf{+14.7\,[+0.9, +28.4]} \\
\midrule
Overall & +1.9\,[-9.4, +13.2] & \textbf{+14.3\,[+5.8, +22.7]} & \textbf{+12.4\,[+9.5, +15.3]} \\
\bottomrule
\end{tabular}
\end{adjustbox}
\end{subtable}
\end{table*}

\section{Detailed analysis: results by category}
\label{sec:catdelta}

Tables~\ref{tab:genevalcat}, \ref{tab:dpgcat} and~\ref{tab:compbench} break the results of Section~\ref{sec:results} down by category on GenEval, DPG-Bench and T2I-CompBench++: the scores after 100 iterations (a) and the paired differences in nAULC (b).

SPROUT reaches the highest score in every DPG-Bench category and in seven of the eight T2I-CompBench++ categories after 100 iterations (Tables~\ref{tab:dpgcat}a and~\ref{tab:compbench}a). On GenEval, it gains most on Two object, which asks for both named objects: from $0.39$ for the base model to $0.58$, against $0.51$ for uniform sampling and $0.50$ for CGPO (Table~\ref{tab:genevalcat}a). A likely reason is the training reward. A Position or Attribute prompt has two requirements, and few images meet both ($5.5\%$ to $8.2\%$ of the training images of these prompts), so the reward of these prompts mostly shows whether the named objects appear. SPROUT favors prompts whose images receive different rewards (Appendix~\ref{sec:informative}), so it trains more on images with and without the named objects. Position remains hard for every method, since it also requires the correct placement; still, SPROUT reaches the highest Position and Attribute scores ($0.08$ and $0.12$). The held-out benchmarks show the same pattern: SPROUT's gains over uniform sampling with intervals that exclude zero are on categories about objects, namely Entity and Relation on DPG-Bench and Shape and 3D-Spatial on T2I-CompBench++. On Non-Spatial, which tests relations such as actions that the training reward does not score, all methods stay within $1\times10^{-3}$ of one another in nAULC.

\begin{table*}[t]
\centering
\small
\caption{DPG-Bench results by category of models trained on GenEval.}
\label{tab:dpgcat}
\begin{subtable}{\linewidth}
\centering
\caption{Score after 100 iterations, mean over three seeds; best results in bold.}
\begin{adjustbox}{max width=\linewidth}\begin{tabular}{lcccccc}
\toprule
Method & Global & Entity & Attribute & Relation & Other & Overall \\
\midrule
Base model & 80.24 & 72.18 & 72.78 & 80.39 & 56.80 & 63.44 \\
Uniform & 82.07 & 73.35 & 73.56 & 80.46 & 56.67 & 64.72 \\
CGPO & 80.95 & 74.51 & 74.28 & 81.90 & 55.87 & 65.46 \\
SPROUT & \textbf{82.17} & \textbf{75.57} & \textbf{75.13} & \textbf{83.57} & \textbf{58.93} & \textbf{67.04} \\
\bottomrule
\end{tabular}
\end{adjustbox}
\end{subtable}
\par\medskip
\begin{subtable}{\linewidth}
\centering
\caption{Differences in nAULC (DPG-Bench points), paired on seed, with bold intervals excluding zero.}
\begin{adjustbox}{max width=\linewidth}\begin{tabular}{lccc}
\toprule
 & \multicolumn{2}{c}{vs Uniform} & vs CGPO \\
\cmidrule(lr){2-3}\cmidrule(lr){4-4}
Category & CGPO & SPROUT & SPROUT \\
\midrule
Global & -0.78\,[-2.17, +0.61] & +0.48\,[-0.78, +1.73] & \textbf{+1.26\,[+0.53, +1.98]} \\
Entity & +0.45\,[-1.56, +2.45] & \textbf{+1.38\,[+0.02, +2.75]} & \textbf{+0.93\,[+0.22, +1.65]} \\
Attribute & -0.05\,[-1.76, +1.67] & +0.68\,[-0.17, +1.54] & +0.73\,[-0.61, +2.07] \\
Relation & +0.56\,[-0.51, +1.62] & \textbf{+1.62\,[+1.29, +1.94]} & \textbf{+1.06\,[+0.26, +1.87]} \\
Other & -0.91\,[-2.95, +1.14] & -0.01\,[-4.23, +4.21] & +0.89\,[-4.17, +5.96] \\
\midrule
Overall & +0.06\,[-1.33, +1.44] & +1.22\,[-0.35, +2.79] & \textbf{+1.16\,[+0.96, +1.37]} \\
\bottomrule
\end{tabular}
\end{adjustbox}
\end{subtable}
\end{table*}

\begin{table*}[t]
\centering
\small
\caption{T2I-CompBench++ results by category of models trained on GenEval.}
\label{tab:compbench}
\begin{subtable}{\linewidth}
\centering
\caption{Score after 100 iterations, mean over three seeds; best results in bold.}
\begin{adjustbox}{max width=\linewidth}\begin{tabular}{lcccccccc}
\toprule
Method & Color & Shape & Texture & 2D-Spatial & 3D-Spatial & Numeracy & Non-Spatial & Complex \\
\midrule
Base model & 0.3744 & 0.3687 & 0.4184 & 0.1059 & 0.3026 & 0.4495 & 0.3108 & 0.3016 \\
Uniform & 0.4595 & 0.4028 & 0.4853 & 0.1612 & 0.3289 & 0.4773 & \textbf{0.3146} & 0.3110 \\
CGPO & 0.4679 & 0.4012 & 0.4796 & 0.1625 & 0.3345 & 0.4764 & 0.3141 & 0.3107 \\
SPROUT & \textbf{0.5111} & \textbf{0.4378} & \textbf{0.5363} & \textbf{0.1830} & \textbf{0.3520} & \textbf{0.4859} & 0.3142 & \textbf{0.3193} \\
\bottomrule
\end{tabular}
\end{adjustbox}
\end{subtable}
\par\medskip
\begin{subtable}{\linewidth}
\centering
\caption{Differences in nAULC by category and for the mean of the eight categories ($\times10^{-3}$), paired on seed, with bold intervals excluding zero.}
\begin{adjustbox}{max width=\linewidth}\begin{tabular}{lccc}
\toprule
 & \multicolumn{2}{c}{vs Uniform} & vs CGPO \\
\cmidrule(lr){2-3}\cmidrule(lr){4-4}
Category & CGPO & SPROUT & SPROUT \\
\midrule
Color & +3.1\,[-37.7, +44.0] & +31.1\,[-7.5, +69.7] & \textbf{+28.0\,[+15.6, +40.4]} \\
Shape & -2.0\,[-16.9, +12.9] & \textbf{+21.3\,[+8.3, +34.2]} & \textbf{+23.2\,[+19.3, +27.2]} \\
Texture & -2.2\,[-34.7, +30.3] & +30.8\,[-5.6, +67.2] & +33.0\,[-2.1, +68.2] \\
2D-Spatial & -0.4\,[-5.8, +5.1] & +12.6\,[-0.5, +25.6] & \textbf{+12.9\,[+4.4, +21.4]} \\
3D-Spatial & -0.4\,[-7.9, +7.1] & \textbf{+12.4\,[+7.1, +17.6]} & \textbf{+12.8\,[+5.3, +20.2]} \\
Numeracy & -2.4\,[-9.5, +4.8] & +5.4\,[-12.6, +23.3] & +7.7\,[-9.8, +25.2] \\
Non-Spatial & -0.7\,[-2.5, +1.1] & -0.8\,[-3.0, +1.4] & -0.1\,[-2.0, +1.8] \\
Complex & -0.6\,[-8.3, +7.0] & +5.0\,[-6.7, +16.7] & \textbf{+5.6\,[+0.4, +10.9]} \\
\midrule
Mean & -0.7\,[-13.2, +11.8] & \textbf{+14.7\,[+0.4, +29.0]} & \textbf{+15.4\,[+8.1, +22.7]} \\
\bottomrule
\end{tabular}
\end{adjustbox}
\end{subtable}
\end{table*}

\begin{figure*}[t]
\centering
\small
\includegraphics[width=\linewidth]{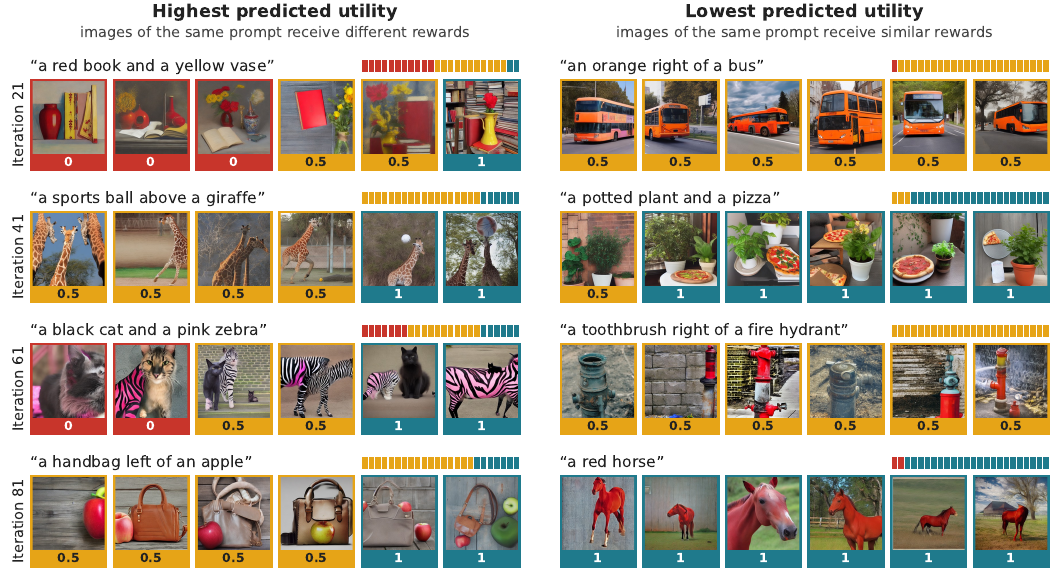}
\caption{Prompts that the utility model of SPROUT ranks highest (left) and lowest (right) among all pool prompts at four iterations, with six of the $24$ generated images for each, picked at a fixed interval after sorting by reward. The frame color and the number under each image give its training reward ($0$, $0.5$, or $1$); the strip beside each prompt shows the rewards of all $24$ images, sorted. The training reward is the fraction of the prompt's requirements that the GenEval verifier finds satisfied, so $1$ means that every requirement is met, $0$ that none is, and $0.5$ that half of them are. Images of a highest-ranked prompt receive different rewards, whereas images of a lowest-ranked prompt receive similar rewards (statistics in Appendix~\ref{sec:informative}).}
\label{fig:informative}
\end{figure*}

\section{Detailed analysis: prompts ranked by the utility model}
\label{sec:informative}

Figure~\ref{fig:informative} shows which prompts SPROUT favors and which it avoids. At iterations 21, 41, 61 and 81 of each of the three seeds, we take the twelve prompts with the highest and the twelve with the lowest predicted utility, generate $24$ images for each from the model of the preceding iteration with the training sampler settings, and score them with the training reward; the figure shows one selected prompt per iteration from each set. Over the $144$ prompts of each set, the within-prompt standard deviation of the reward is $0.250$ for the highest-ranked and $0.180$ for the lowest-ranked prompts (difference $+0.070$, 95\% interval $[0.005, 0.136]$ over three seeds). Prompts drawn uniformly from the pool lie in between, with a within-prompt standard deviation of $0.227$. The utility model is fit on update norms, not on rewards or their variance, yet the prompts it ranks highest are those whose images the reward separates, which is the within-group contrast that the preference objective learns from (Section~\ref{sec:objective}): images whose rewards differ receive nonzero advantages, whereas a group with equal rewards contributes only through the penalty to the reference model.

\begin{table*}[t]
\centering
\small
\caption{Utility, label history and batch selection, against uniform sampling on GenEval.}
\label{tab:ablation}
\begin{adjustbox}{max width=\linewidth}\begin{tabular}{lllcc}
\toprule
Utility & History & Batch selection & $\Delta$nAULC vs Uniform ($\times10^{-3}$) & Reduction factor \\
\midrule
Variance & previous & top-$B$ & +3.1\,[-9.4, +15.5] & 1.43$\times$ \\
Variance & latest & top-$B$ & +0.1\,[-26.6, +26.8] & -- \\
Variance & all & top-$B$ & -20.1\,[-141.9, +101.7] & -- \\
Variance & previous & rejection & +8.2\,[-3.9, +20.2] & 1.63$\times$ \\
Variance & all & rejection & \textbf{+9.7\,[+5.7, +13.8]} & 1.80$\times$ \\
Update norm & all & top-$B$ & +8.0\,[-15.3, +31.2] & 1.63$\times$ \\
Update norm & all & rejection & \textbf{+14.3\,[+5.8, +22.7]} & 2.27$\times$ \\
\bottomrule
\end{tabular}
\end{adjustbox}
\end{table*}

\section{Ablation: label history and batch selection}
\label{sec:ablation}

We study three \emph{label histories}, the subsets of past records that the utility model is fit on: every record so far (\emph{all}), the most recent record of each prompt visited so far (\emph{latest}), or the records of the preceding iteration only (\emph{previous}).
Table~\ref{tab:ablation} varies one component at a time, and batch selection matters most. With predicted variance, top-$B$ gains little over uniform sampling under previous-iteration or latest-record labels and collapses on one seed under all-history labels (GenEval below $0.001$ at iteration 100), whereas rejection sampling on the same predictions never collapses, keeps a gain under both label histories tested, and beats top-$B$ on every seed with previous-iteration labels ($+5.1\times10^{-3}$, 95\% interval $[+4.5, +5.7]$).

Top-$B$ does not fall behind for lack of informative groups. Counting a group as informative when its reward variance exceeds uniform sampling's median at the same iteration, top-$B$ trains on many more of them than uniform sampling ($65\%$ with previous-iteration and $87\%$ with latest-record labels, against $52\%$), yet with latest-record labels it neither gains in nAULC ($+0.1\times10^{-3}$) nor reaches uniform sampling's best score within $100$ iterations, while SPROUT, with the largest reduction factor, raises the share only to $56\%$. What top-$B$ gives up is coverage: its trained prompts lie closer to their nearest trained neighbor in CLIP~\citep{radford2021clip} embedding space, with a mean cosine similarity (the average nearest embedding similarity of UniFL~\citep{zhang2024unifl}) of $0.856$ against $0.803$ for uniform sampling (difference $+0.053$, 95\% interval $[+0.023, +0.082]$), while rejection sampling stays at uniform sampling's level for both utilities.

The update norm shows the same concentration without a collapse. Top-$B$ trains on only $279$ to $461$ distinct prompts over $100$ iterations, against about $790$ for rejection sampling, repeats a single prompt up to $78$ times, and gains less over uniform sampling ($8.0\times10^{-3}$ against $14.3\times10^{-3}$ in nAULC and $1.63\times$ against $2.27\times$ in reduction factor), although the difference between the two rules has an interval that includes zero.

\begin{table*}[t]
\centering
\small
\caption{GenEval results with $B=16$ prompts per iteration. Columns as in Table~\ref{tab:main}. A dash under reduction factor means the method does not reach uniform sampling's best score within $100$ iterations.}
\label{tab:batch16}
\begin{adjustbox}{max width=\linewidth}\begin{tabular}{lccccc}
\toprule
Method & GenEval@100 & nAULC & $\Delta$nAULC vs Uniform & $\Delta$nAULC vs CGPO & Reduction factor \\
 & & & ($\times10^{-3}$) & ($\times10^{-3}$) & \\
\midrule
Base model & 0.426 & -- & -- & -- & -- \\
Uniform & 0.490\,$\pm$\,0.027 & 0.469\,$\pm$\,0.011 & -- & -- & 1.00$\times$ \\
CGPO~\cite{Li_2026_CVPR} & 0.487\,$\pm$\,0.064 & 0.462\,$\pm$\,0.021 & -7.2\,[-18.3, +3.9] & -- & -- \\
SPROUT w/ variance & 0.505\,$\pm$\,0.018 & 0.479\,$\pm$\,0.005 & +9.5\,[-6.1, +25.2] & +16.7\,[-7.3, +40.7] & 1.75$\times$ \\
SPROUT & 0.500\,$\pm$\,0.014 & 0.477\,$\pm$\,0.021 & +7.8\,[-11.7, +27.3] & +15.0\,[-1.7, +31.7] & 1.70$\times$ \\
\bottomrule
\end{tabular}
\end{adjustbox}
\end{table*}

\section{Additional experimental setting: larger batches}
\label{sec:batch16}

\begin{figure}[t]
\centering
\includegraphics[width=\linewidth]{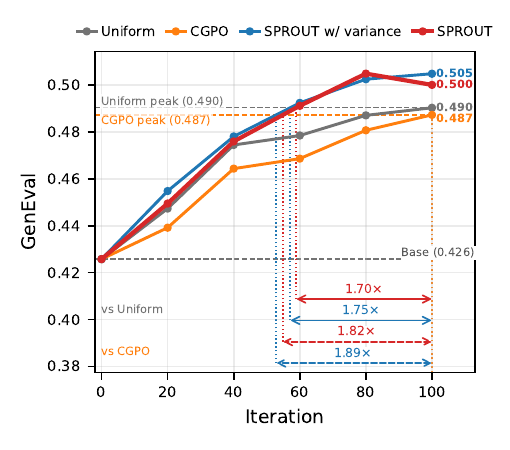}
\caption{GenEval $B=16$ learning curves over training, mean of three seeds. Arrows mark the reduction factor relative to the best scores of uniform sampling (solid) and CGPO (dashed).}
\label{fig:batch16}
\end{figure}

Table~\ref{tab:batch16} and Figure~\ref{fig:batch16} repeat the comparison with $B=16$ prompts per iteration. Uniform sampling benefits from the larger batch ($0.490$ after $100$ iterations against $0.470$ at $B=8$), yet SPROUT and SPROUT w/ variance still lie above it in nAULC on every seed and reach its final score with a reduction factor of about $1.7\times$, while CGPO does not reach it on the seed mean. With three seeds the GenEval intervals include zero, but the held-out reward models and DPG-Bench at iteration 100 (Table~\ref{tab:batch16transfer}) favor both on every seed, significantly so for SPROUT w/ variance under PickScore and ImageReward and on DPG-Bench ($+1.83$ points). On T2I-CompBench++, both raise the mean over the eight categories on every seed, by $+19.3\times10^{-3}$ for SPROUT w/ variance and $+14.4\times10^{-3}$ for SPROUT, with intervals that exclude zero, while the interval of CGPO includes zero. Training also stays stable, with no run of either selection method falling more than $0.014$ below its best checkpoint. Appendix~\ref{sec:predabl} compares other predictors at this batch size.

\begin{table*}[t]
\centering
\small
\caption{Held-out evaluations with $B=16$: difference from uniform sampling in nAULC over iterations $0$--$100$ under the held-out reward models (scaled as in Table~\ref{tab:judges}), and in score at iteration 100 on DPG-Bench (points) and T2I-CompBench++ (mean of the eight categories, $\times10^{-3}$), paired on seed.}
\label{tab:batch16transfer}
\begin{adjustbox}{max width=\linewidth}\begin{tabular}{lccc}
\toprule
 & CGPO~\cite{Li_2026_CVPR} & SPROUT w/ variance & SPROUT \\
\midrule
\multicolumn{4}{l}{$\Delta$nAULC vs Uniform, iterations 0--100} \\
PickScore & +0.4\,[-3.5, +4.3] & \textbf{+4.0\,[+0.2, +7.7]} & +3.4\,[-3.0, +9.7] \\
HPS v2.1 & -0.8\,[-6.8, +5.1] & +3.7\,[-0.3, +7.7] & +2.3\,[-0.9, +5.5] \\
Aesthetic & +1.4\,[-5.1, +7.9] & +3.7\,[-2.1, +9.5] & +3.6\,[-0.9, +8.1] \\
ImageReward & -3.7\,[-16.0, +8.5] & \textbf{+6.3\,[+2.9, +9.6]} & +3.3\,[-3.7, +10.3] \\
\midrule
\multicolumn{4}{l}{$\Delta$score vs Uniform at iteration 100} \\
DPG-Bench (points) & -0.87\,[-5.63, +3.90] & \textbf{+1.83\,[+0.98, +2.67]} & +1.06\,[-1.20, +3.31] \\
T2I-CompBench++ ($\times10^{-3}$) & -9.8\,[-51.6, +32.0] & \textbf{+19.3\,[+4.5, +34.0]} & \textbf{+14.4\,[+6.3, +22.5]} \\
\bottomrule
\end{tabular}
\end{adjustbox}
\end{table*}

\begin{table*}[t]
\centering
\small
\caption{Predictors at $B=16$ prompts per iteration: GenEval after 100 iterations and difference in nAULC from uniform sampling, over three seeds. Ridge regression with $\alpha=10$ (rows 1 and 3) ranks prompts worse offline than the alternatives (rows 2 and 4; Spearman correlations in the text), yet attains higher mean GenEval and nAULC.}
\label{tab:batch16pred}
\begin{adjustbox}{max width=\linewidth}\begin{tabular}{llccc}
\toprule
Target & Predictor & GenEval@100 & nAULC & $\Delta$nAULC vs Uniform ($\times10^{-3}$) \\
\midrule
Variance & Ridge, $\alpha=10$ & 0.505\,$\pm$\,0.018 & 0.479\,$\pm$\,0.005 & +9.5\,[-6.1, +25.2] \\
Variance & Ridge, $\alpha=1000$ & 0.481\,$\pm$\,0.112 & 0.473\,$\pm$\,0.016 & +3.5\,[-11.7, +18.8] \\
Update norm & Ridge, $\alpha=10$ & 0.500\,$\pm$\,0.014 & 0.477\,$\pm$\,0.021 & +7.8\,[-11.7, +27.3] \\
Update norm & Fine-tuned CLIP & 0.403\,$\pm$\,0.413 & 0.466\,$\pm$\,0.047 & -3.5\,[-60.4, +53.4] \\
\bottomrule
\end{tabular}
\end{adjustbox}
\end{table*}

\section{Ablation: predictors on larger batches}
\label{sec:predabl}

Table~\ref{tab:batch16pred} asks whether a stronger predictor helps, with every other setting as in Table~\ref{tab:batch16}. Against the ridge regression with $\alpha=10$ used throughout the paper, we test ridge with $\alpha=1000$ for the variance utility and, for the update norm, the CLIP ViT-L/14 text encoder with a new $768$--$256$--$1$ head, fine-tuned at every iteration with one pass over all labels so far (learning rate $10^{-5}$ for the encoder and $3\times10^{-4}$ for the head). Offline, fitting on earlier iterations of $B=8$ runs and predicting the prompts trained at the next, both rank prompts better than ridge with $\alpha=10$ (Spearman $0.224$ against $0.171$ for the variance utility and $0.215$ against $0.077$ for the update norm). This ranking is almost entirely of prompts without a label of their own: $98.5\%$ of the prompts that SPROUT trains on with $B=8$ had never been trained on earlier in the same run, and on these prompts the Spearman correlation between the utility model's prediction at selection time and the label is $0.027$ in iterations 2--20 and $0.166$ in iterations 81--100. In training they are no better: both lead uniform sampling at iterations 60 and 80 on the seed mean, but one run of each degrades once the policy drifts from the reference model, the worst falling from $0.501$ at iteration 80 to $0.212$ at 100. Ridge with $\alpha=10$ keeps the largest nAULC gain for both utilities and none of its runs degrades, so a better offline ranking does not translate into better training here.

\begin{table*}[t]
\centering
\small
\caption{GenEval over 200 iterations, with the learning rate decayed after iteration 100. nAULC and reduction factor are computed over iterations $0$--$200$; other columns as in Table~\ref{tab:main}.}
\label{tab:decay200}
\begin{adjustbox}{max width=\linewidth}\begin{tabular}{lcccccc}
\toprule
Method & GenEval@100 & GenEval@200 & nAULC & $\Delta$nAULC vs Uniform & $\Delta$nAULC vs CGPO & Reduction factor \\
 & & & & ($\times10^{-3}$) & ($\times10^{-3}$) & \\
\midrule
Base model & 0.426 & 0.426 & -- & -- & -- & -- \\
Uniform & 0.470\,$\pm$\,0.051 & 0.497\,$\pm$\,0.028 & 0.470\,$\pm$\,0.023 & -- & -- & 1.00$\times$ \\
CGPO~\cite{Li_2026_CVPR} & 0.475\,$\pm$\,0.015 & 0.472\,$\pm$\,0.146 & 0.464\,$\pm$\,0.056 & -5.8\,[-43.3, +31.6] & -- & -- \\
SPROUT w/ variance & 0.493\,$\pm$\,0.032 & 0.504\,$\pm$\,0.042 & 0.483\,$\pm$\,0.030 & +12.8\,[-23.3, +49.0] & +18.7\,[-54.5, +91.9] & 1.81$\times$ \\
SPROUT & 0.483\,$\pm$\,0.026 & 0.507\,$\pm$\,0.020 & 0.483\,$\pm$\,0.006 & +12.8\,[-5.1, +30.6] & +18.6\,[-31.8, +69.1] & 1.50$\times$ \\
\bottomrule
\end{tabular}
\end{adjustbox}
\end{table*}

\section{Additional experimental setting: training beyond 100 iterations}
\label{sec:beyond}

\begin{figure}[t]
\centering
\includegraphics[width=\linewidth]{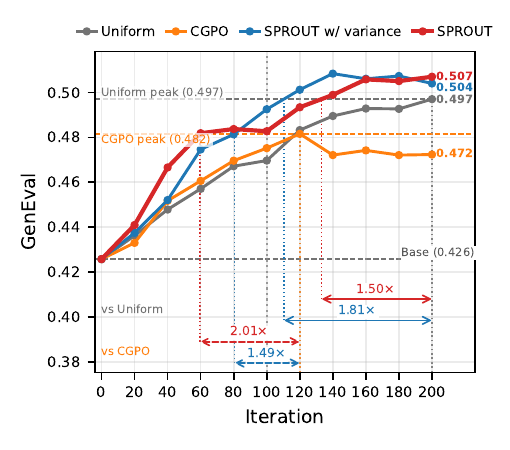}
\caption{GenEval learning curves over 200 iterations with the learning rate decayed after iteration 100, mean of three seeds. Arrows mark the reduction factor relative to the best scores of uniform sampling (solid) and CGPO (dashed).}
\label{fig:curves200}
\end{figure}

To see whether the gains persist, we resume each $B=8$ run from its state at iteration 100 and continue to iteration 200. At a constant learning rate, prompt selection becomes unstable: three of the six runs of SPROUT and SPROUT w/ variance collapse, falling more than $0.02$ below their own best at two consecutive checkpoints, against neither of the two uniform-sampling runs extended. Every method in Table~\ref{tab:decay200} and Figure~\ref{fig:curves200} therefore decays the learning rate after iteration 100, by a cosine to $0.3\times$ at iteration 120 and a second cosine to zero at iteration 200. Under this schedule none of our runs collapses, while CGPO collapses on one seed, and SPROUT w/ variance and SPROUT keep their lead over uniform sampling, at $0.504$ and $0.507$ against $0.497$ at iteration 200. Their nAULC gains of $12.8\times10^{-3}$ have intervals that include zero, with SPROUT ahead on all three seeds and SPROUT w/ variance on two. The held-out reward models and the other benchmarks agree (Table~\ref{tab:decay200transfer}): both lie above uniform sampling on the seed mean under all four held-out reward models and on DPG-Bench and T2I-CompBench++ at iteration 200, with intervals that include zero, and SPROUT improves both benchmarks on all three seeds. CGPO, with one collapsed seed, falls below uniform sampling under three of the four reward models and on both benchmarks.

\begin{table*}[t]
\centering
\small
\caption{Held-out evaluations over 200 iterations: difference from uniform sampling in nAULC over iterations $0$--$200$ under the held-out reward models (scaled as in Table~\ref{tab:judges}), and in score at iteration 200 on DPG-Bench (points) and T2I-CompBench++ (mean of the eight categories, $\times10^{-3}$), paired on seed.}
\label{tab:decay200transfer}
\begin{adjustbox}{max width=\linewidth}\begin{tabular}{lccc}
\toprule
 & CGPO~\cite{Li_2026_CVPR} & SPROUT w/ variance & SPROUT \\
\midrule
\multicolumn{4}{l}{$\Delta$nAULC vs Uniform, iterations 0--200} \\
PickScore & -1.3\,[-18.2, +15.6] & +3.6\,[-19.0, +26.2] & +4.9\,[-12.6, +22.4] \\
HPS v2.1 & -2.9\,[-11.9, +6.0] & +4.2\,[-5.4, +13.8] & +4.7\,[-5.8, +15.2] \\
Aesthetic & +2.8\,[-4.8, +10.3] & +3.4\,[-2.1, +8.8] & +3.5\,[-0.9, +8.0] \\
ImageReward & -4.7\,[-25.2, +15.7] & +9.3\,[-13.9, +32.5] & +8.5\,[-5.1, +22.2] \\
\midrule
\multicolumn{4}{l}{$\Delta$score vs Uniform at iteration 200} \\
DPG-Bench (points) & -2.95\,[-17.51, +11.61] & +0.49\,[-4.47, +5.45] & +1.27\,[-0.85, +3.39] \\
T2I-CompBench++ ($\times10^{-3}$) & -24.1\,[-121.0, +72.8] & +10.2\,[-25.8, +46.2] & +13.7\,[-8.4, +35.8] \\
\bottomrule
\end{tabular}
\end{adjustbox}
\end{table*}

\end{document}